\documentclass{article}

\usepackage{arxiv}

\usepackage[utf8]{inputenc} 
\usepackage[T1]{fontenc}    
\usepackage{hyperref}       
\usepackage{url}            
\usepackage{booktabs}       
\usepackage{amsfonts}       
\usepackage{nicefrac}       
\usepackage{microtype}      
\usepackage{lipsum}
\usepackage{caption}
\usepackage[title]{appendix}
\usepackage{multirow}
\usepackage{changepage}
\usepackage{multicol}
\usepackage{tabularx}
\usepackage{pifont}
\usepackage{ragged2e}
\usepackage{subcaption}
\usepackage{graphicx}
\usepackage{float}
\usepackage[dvipsnames]{xcolor}
\graphicspath{ {./images/} }
\usepackage{tocloft}

\newcommand{\cmark}{\ding{51}}
\newcommand{\xmark}{\ding{55}}

\title{Agentic LLM Framework for Adaptive Decision Discourse}

\author{
 Antoine Dolant \\
  Department of Civil and Environmental Engineering\\
  Grainger College of Engineering \\
  University of Illinois Urbana-Champaign\\
  Urbana, IL 61801 \\
  \texttt{adolant2@illinois.edu} \\
   \And
 Praveen Kumar \\
  Civil and Environmental Engineering\\
  Grainger College of Engineering \\
  Prairie Research Institute \\
  University of Illinois Urbana-Champaign\\
  Urbana, IL 61801 \\
  \texttt{kumar1@illinois.edu} \\
}

\hypersetup{
 colorlinks=true,
 citecolor=NavyBlue,
 linkcolor=BrickRed
 }

\begin{document}
\maketitle
\begin{abstract}
Effective decision-making in complex systems requires synthesizing diverse perspectives to address multifaceted challenges under uncertainty. This study introduces an agentic Large Language Models (LLMs) framework for simulating decision discourse — the deliberative process through which actionable strategies are collaboratively developed. Unlike traditional decision-support tools, this framework simulates diverse stakeholder personas, each bringing unique priorities, expertise and value-driven reasoning to a dialogue that emphasizes trade-off exploration in a self-governed assembly. We present explorative results fostering robust and equitable recommendations, with two use cases: first, our framework simulates a response to the floods that occurred on July 2025 in Texas; second, a hypothetical extreme flooding in a Midwestern township under varying forecasting uncertainty. Recommendations made balance competing priorities considered through social, economic and environmental dimensions, setting a foundation for scalable and context-aware recommendations and transforming how decisions for real-world high-stake scenarios can be approached in digital environments. This research explores novel and alternate routes leveraging agentic LLMs for adaptive, collaborative, and equitable recommendations, with implications across domains where uncertainty and complexity converge.
\end{abstract}

\section{A Case for LLMs in Human Discourse}
\label{titles:introcomplexity}

Human decision-making unfolds within a landscape marked by uncertainty and risk, complexity, and competing priorities. This is particularly true in scenarios requiring immediate yet informed responses to multifaceted challenges, where the stakes encompass social, economic, and environmental dimensions. Natural hazard scenarios, such as those posed by extreme weather events, underscore this challenge. Climate-driven disasters like floods, wildfires, and hurricanes are shaped by dynamic, coupled human-natural systems (CHANS). These systems involve a multitude of interacting biophysical and social processes that can amplify cascading consequences over varying spatial and temporal scales. Decision-making in such contexts must account for resource limitations, socio-economic vulnerabilities, and the inherent unpredictability of impending events \cite{clarke2018}. In this context, decision support is described as “the intersection of data provision, expert knowledge, and human decision making at a range of scales from the individual to the organization and institution” \cite{jones2014}. Whether addressing natural disasters, public health crises, or systemic societal inequities, decisions often depend on limited information, interdisciplinary expertise, and the need to weigh short and long-term consequences. Despite decades of advancements in decision-support systems, the human dimension of decision-making — the intuitive and value-driven processes of individuals and communities — still remains difficult to quantify and incorporate into actionable frameworks.

Existing decision frameworks often rely on mathematical and physical models that inadequately capture the full spectrum of the multi-dimensional complexity of the problem at hand. Existing examples from behavioral economics include the beta-delta model \cite{laibson1997}, disinterested distributional preferences \cite{adler2000} and loss aversion model \cite{tversky1991}, all of which offer mathematical representation of aspects of human preferences in making decisions. These approaches have proven their value with regards to purchasing, selling and simple decision-making modeling, yet these models are too coarse to fully capture the intricacy of complex problems such as preparedness to uncertain natural hazards. They often fail to fold the human dimension into state of the  art approaches as a result of its inherent complexity, which is hardly expressed by mathematical and physical-based models. Such limitations partially explain the under-representation of human decision-making in mitigation and adaptation efforts to extreme events assessment. Notably, social equity and justice has been correlated to socio-economic principles and considered as a major challenge in mitigation for decades, yet was only truly considered recently \cite{ikeme2003}. Nonetheless, numerous interacting processes contribute to CHANS’ complex functioning, hampering the ability to address the challenges associated with strategic planning, preparedness and disaster recovery \cite{townend2023, mach2023}.

Uncertain decision-making is regarded as complex as a result of the numerous cascading consequences that could stem from an initial action, hence strongly encoding causality. Anticipating consequences require parallel and pluridisciplinary thinking, so as to causally infer potential effects from observations and interventions. Such parallel thinking is easier represented by multi-agent systems, for which specialized objectives embedded in various disciplines can we distributed across different agents. The multiplication of model instances — such as specialized agents optimizing for different objectives — is referred to as horizontal scaling. This contrasts with monolithic Large Language Models (LLMs), that have scaled vertically over the past few years \cite{kaplan2020}, favoring single-actor performance over collaborative behaviors \cite{feng2025mas}. Additionally, LLM agents have shown emerging social behaviors \cite{park2023, riedl2025, feng2025llmdrools}, as well as causal capabilities encoded in language \cite{liu2019}, aligning with human decision-making objectives. Crystallizing the claim that “experience is very important for intelligent agents to understand causality”, LLMs were demonstrated to be state of the art in finding semantic causal relationships between sentences \cite{liu2019, lazaridou2017}. Their semantic causal capabilities and their massive human-produced knowledge combined, LLMs are often described as world model representations \cite{gurnee2024}. In addition to agentic emerging social behaviors and parallel processing of distinct domain objectives, we posit that agentic LLM systems are far more adapted for decision making under uncertainty, compared to monolithic models.

In this study, we propose and evaluate a multi-agent framework powered by Large Language Models, designed to simulate and enhance decision-making processes. Through this approach, we aim to organically develop the interactive potential of LLM agents as a tool for discourse and synthesis to explore decision making recommendations. This framework augments the capacity of decision-makers to navigate complexity by providing arguments drawn upon prior knowledge, knowhow and outcomes. Moreover, it lays a foundation for scalable, adaptive, and context-aware recommendations to emergent challenges. Our findings highlight the potential of LLM-based multi-agent systems to produce actionable strategies and highlight adaptive decision pathways to evolving risks in integrative applications. Outside the scope of disaster resilience, we present broader impacts of LLM-powered agentic decision discourse. Broader impacts include multi-agent driven synergistic behaviors, dynamical system modeling using LLMs, and information theoretic approaches to understand agent interactions.

This manuscript is organized as follows: we present the foundation for a multi-agent LLM framework built to capture the interactions of decision-makers during the process of challenge assessment. Motivations for such a structural design are detailed and supported by information theoretic thinking. Then we present an assessment of conversation mechanisms and their implementation into LLM-powered agents augmented with prompt engineering techniques, as well as the requirements for the operational framework to support discourse between agents.
With two use cases of impending hazards, we present promising perspectives of the framework's ability to navigate competing priorities, integrate interdisciplinary insights, and evaluate adaptive strategies.
We highlight a dynamic relationship between resulting strategies and uncertainty.
The results underscore the capabilities of such a framework to serve as a digital twin for human decision discourse, fostering scalable and context-aware recommendations.

\section{Multi-Agent Framework for Decision Discourse}
Transformer-class models were introduced in 2017 \cite{vaswani2023} and a few years forward saw the birth of ChatGPT, a conversational LLM application that is still the fastest-growing consumer software application ever produced \cite{wikipedia1}. LLMs are trained on massive text corpora that concentrate human knowledge to perform next-word prediction and iteratively generate an infinite amount of text. They have proven to outperform any type of Machine Learning or non-Machine Learning model in natural language processing (NLP) tasks \cite{rostam2024}. Common LLM usage includes questioning and answering, summarizing, translation, creative writing and semantic tasks for which they secured their place at the top of the leaderboard  \cite{niimi2024}. More recently, LLM reasoning has gained considerable representation in computer science and AI research. Furthermore, LLMs have shown potential for causal discovery and observational causality inference, although the full extent of their capacity to perform advanced causal analysis is an active research and discussion \cite{jin2024cladder, jin2024, kiciman2023, zevcevic2023}. The exhaustiveness of the training dataset enables them to simulate nuanced discourse, evaluate hypothetical scenarios, and dynamically adapt to contextual shifts. Unlike traditional decision-making tools, LLMs can engage with the rich interplay of human perspectives, integrating expertise across disciplines and negotiating trade-offs through collaborative human-like conversations. As such, they represent a promising candidate for addressing the intricacies of decision-making under uncertainty. These precise characteristics have brought LLMs to the foreground of human-like conversational models as their capabilities encapsulate nuanced opinions and approaches, hypothetical situation evaluation, and dynamic adaptation to situational contexts. Notwithstanding the paradigm shift from monolithic infrastructures to distributed systems that the world has experienced in the past decades \cite{mosleh2018}, LLMs are trained and built as unique and enormous single instance models. The popularization of agentic framework has gained a lot of representation through the course of 2025, although the base models are still monolithic. Drawing a parallel with CHANS—that are complex systems with many interacting parts, we posit that multi-agent LLM systems manifest a stronger topological relevance when compared to single-agent or monolithic implementations.

\subsection{Framework Outline}
\label{titles:frameworkoutline}
Recent years have seen a rapidly growing literature including decision-making focused LLM \cite{zhao2023, hao2023, zhou2024}, agentic LLM frameworks \cite{dong2024, campedelli2024, chang2024socra, zhou2024, talebirad2023}, and agentic conversational frameworks \cite{park2023, chan2023, chang2024evince, cho2024, estornell2024}. In this landscape of rapidly evolving models, our focus is to present a first exploration of agentic LLM decision discourse, that sets multiple simulated stakeholder providing decision support in the face of uncertainty. Our agents are built with the Google ADK development kit, that provides customizable agent workflows in Python. Google ADK allows to write generic agent-building code, to which user-defined behaviors and controlled flow instructions are added, resulting in the framework we present over the next sections. Additional details about the ADK implementation, and the technical framework structure are available in Appendix \ref{titles:additionalframework}.

\begin{figure}
\centering
\includegraphics[width=\textwidth]{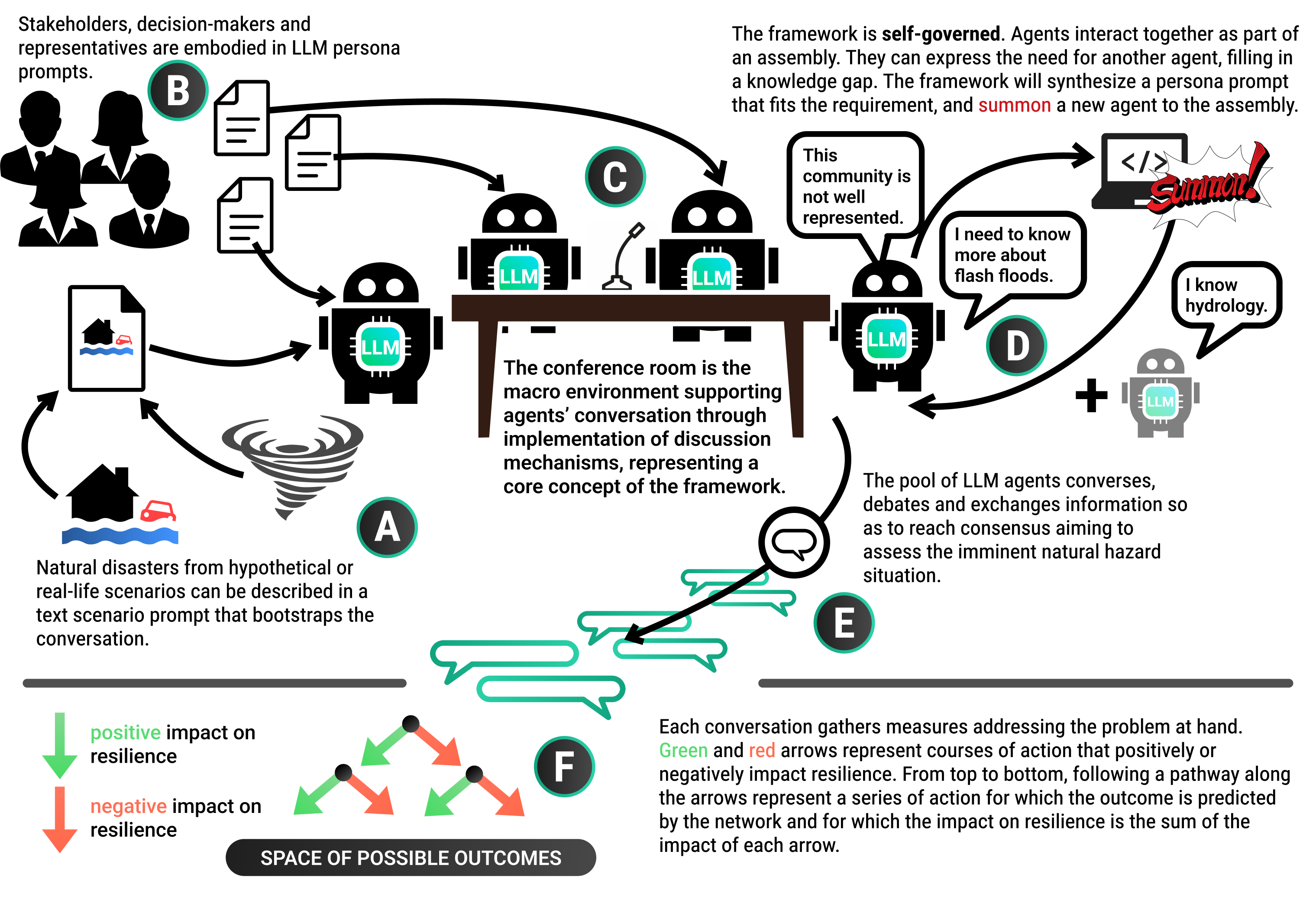}
\caption{Illustration of the multi-agent decision discourse workflow.   First (A),  a natural disaster threat is selected to generate a scenario that will act as a bootstrap to LLM agents.  Second (B),
initial persona prompts are specified or generated for the starting assembly (for example, composed of a mayor, community advocate, environmental scientist, and moderator for an extreme flooding scenario).  With those two necessary conditions, the assembly (C) starts and iterates for a specified number of runs or until convergence. Each iteration consists of an agent continuing the discussion at time state \textit{t} from the conversation state at time \textit{t-1} as an input. Agents can bring in new participants (D) if they judge that there is an imbalance in the skills and knowledge domains represented within the assembly. Each  execution results in the generation of a conversation output (E), from which can be extracted recommended courses of action with their advantages and drawbacks. Actionable measures are colored here as green or red (F) depending on their positive or negative impact on resilience to the situation. The space of possible outcomes represents all realizations of the different action pathways, and can be enriched with additional executions  with slightly varying inputs, evaluating scenarios that are causally defined through counterfactuals.
}
\label{fig:thiswork}
\end{figure}

Our overall framework operates as a simulation, beginning with a scenario prompt that sets a starting point for discussion (see Fig.\ \ref{fig:thiswork}). For the scenario, tasks are generated and ranked from most relevant to least relevant. The number of tasks that are selected for processing is a user-defined parameter. Evaluation of the relevance of a task uses an approximation of Mutual Information between the input scenario and the task. Additional details are presented in Appendix \ref{titles:unfolding}. Similar to conversational AI, agent responses are structured into a message thread, the first message being the scenario prompt. At each round, one agent generates a message, followed by the opportunity to summon another agent, and to assess the current state of the conversation. Both those self-governance mechanisms are triggered at a user-defined frequency, ensuring that the framework remains adaptive and context-aware. As messages are generated, agents engage in iterative discourse, exchange ideas, challenge assumptions, and refine strategies based on feedback. When the number of messages reaches a user-defined limit, discourse stops. The last iteration is followed by the synthesis of results and summary analysis, stating evaluated strategies, their feasibility and plausibility, and advantages and drawbacks. From this final assessment can be extracted actionable recommendations, represented by the green and red arrow graph on Fig. \ref{fig:thiswork}. A more streamlined and comprehensive representation of the framework's operation is presented in Appendix \ref{titles:additionalframework}. In this section, we present the foundations of this conceptual framework, develop the different components of the framework, and argue for the necessity for self-governance.

\subsection{Agentic Prompt Engineering}
\label{titles:agenticpromptengineering}
Agentic LLM designs go hand in hand with prompt engineering techniques, which allow a deep behavioral modification of the model, observed in generated outputs. Prompt engineering is defined as a set of techniques modifying the behavior of the data retrieval mechanism of LLMs through structural patterns: precise choice of words, ordering of instructions, and formatting templates. For instance, a persona prompt pattern constrains the model into impersonating a particular character, so as to enforce a particular behavior. By way of illustration, instructing an LLM to answer as if it was addressing a fourth graders class will impact the vocabulary, sentence construction and logical reasoning employed to formulate the answers to match the audience. Building on those mechanisms, agents can impersonate humans with defined behavioral profiles, character traits, and professional and/or personal objectives, thereby simulating human-like interactions. As a consequence, the accessible knowledge space of the model is restrained, without which LLMs’ behavior would not be consistent with the described persona prompt. For agentic LLM workflows to fully leverage the interactions between agents, these must present specific behaviors. Indeed, recent work has shown that redundant behaviors contributes to a performance increase when objectives and goals align within the agent's persona \cite{riedl2025}. Nevertheless, we posit that the choice of persona is a tradeoff between positive redundancy that contributes to performance, and narrow knowledge representations showing for little overlap, hence reducing opportunities to reconcile competing objectives.

\subsection{Agentic Foundation for Decision Discourse}
\label{titles:conceptualfoundation}

Following the principles described in the previous  section, recent work \cite{talebirad2023} introduces a general framework design that supervises LLM-based agents aspiring to “enhance the performance and capabilities of LLMs by leveraging the power of multi-agent systems”. This top-down initiative aims to pave the way for the creation of more powerful Artificial General Intelligence (AGI) models \cite{talebirad2023}. Central to this goal, agents are seen as executors that work collaboratively towards certain tasks, sometimes requiring specialized behaviors. This approach presents multi-agent LLMs as a more efficient and comprehensive task-solving framework, for which applications vary across field. Although demonstrating increased performance in complex technical tasks, such initiatives underperform in applications that include complex decision making under uncertainty \cite{mosleh2018}.

In comparison, we develop a bottom-up workflow drawing from agentic personas with specialized knowledge, towards the goal of leveraging the semantic capabilities of LLMs. In the context of decision-making under uncertainty — especially in disaster resilience challenges, the richness of human viewpoints and reasoning is hardly captured by non-LLM approaches. Hence, we view agentic LLM systems as a proxy for human decision-making, leveraging collaborative and explorative discourse as means to present competing solution pathways. We model agents as distinct human-inspired personas, each representing a specific stakeholder role (elected officials, domain specialists, etc.) that possesses distinctive deeper values such as risk awareness, tolerance and social equity considerations. We advance that our approach draws interest from discourse between agents whose accessible knowledge is restricted by their persona definition, rather than tool-use and other technical features of agentic systems. This statement reflects on our argument that monolithic LLMs are limited in simulating simultaneous behaviors, as a direct consequence of their topology.

Our framework draws on principles of system thinking, to organically construct a core workflow inspired by real-world societal structures that leverages interactions. Drawing a parallel with such real-world societal structures, our agentic system is designed by drawing inspiration from multiple individuals holding an emergency response meeting. The diversity of characters in this assembly is purposed to dynamically explore recommendation pathways, relevant with different aspects of the uncertainty challenge and representative of different expertise. Such aspects include population safety with considerations regarding social equity, and integrity of vital infrastructures. 

\subsubsection{Structural Design}
\label{titles:structuraldesign}

In this section, we use the concepts previously introduced to define the first necessary requirement of agentic LLM workflow supporting human-like decision-making. This specification will draw heavy inspiration from real-world decision-making contexts, where decisions could be conceptualized as emergent products of human discourse and careful peer evaluation involving consensus. Decision-making necessarily involves a physical or a virtual space, in which interlocutors can meet. Good recommendations “tend to emerge from processes in which people are explicit about their goals; consider a range of alternative options for pursuing their goals; use the best available science to understand the potential consequences of their actions; carefully consider the trade-offs; and contemplate the decision from a wide range of views and vantages” \cite{jones2014}. Similarly, agents need a space to meet, debate and converge to a series of recommendations. As is it mentioned in previous sections, monolithic LLM implementations do not allow for communication between multiple entities, in opposition to Multi-Agent Systems (MAS) whose core functioning articulates around peer interactions. Hence, we create the virtual conference room, that gathers the mechanisms allowing inter-agent communication, and exposes features for LLM agent orchestration. Among those, a shared memory space to store agentic interactions, mechanisms to determine the speaking order, and mechanisms to detect whether new agents should join the room. For simplicity, we refer to conference room-level actions or features as macro-scale, and agent-level as micro-scale. Defined as such, the virtual conference room represents a key and necessary component to enable agentic LLM interactions, without which emergent products of human discourse cannot exist.

\subsubsection{Agent Design}
\label{titles:agentdesign}

Similar to their human counterparts, the key capability necessary for LLM agents to interact is a dialogue mechanism. A typical iteration of dialogue consists of a listening phase, a sensible answer construction, and an expressive phase. Using computer science vocabulary, this translates to input, processing, and output, respectively attributed to sensors, internal logic and actuators in MAS theory \cite{holland1995}. In LLMs, sentence construction is supported by their model parameters which encode the natural language function. Hence, prompts can be assimilated to listening, and the output to the expression. The proximity between LLMs' structure and human discourse is partially caused by the compressed representations of our world that are embedded in LLMs’ inner structures (Section \ref{titles:hallucinations}).

Next, in order to carry out a conversation, agents need memory to remember past interactions. This conversational memory is integrated to web conversational versions of the most popular LLMs (ChatGPT, LLaMa, etc.), and often absent from API versions. More generally, web conversational versions include more user-oriented features such as contextual memory, tool-use and file uploads, while API versions focus on the integration of LLMs into user-defined applications at the expense of aforementioned features. Choosing between the web conversational or the API version is a tradeoff that considers the amount of needed features or the granularity of control needed over LLMs. In other words, at present LLMs are either available as conversational agents in their monolithic form, or as an application programming interface (API) that enables the integration in more flexible applications. The API form being stateless, the implementation of conversational memory and other features is left to developers. To achieve this in an agentic LLM framework, we store the conversation at the macro-scale level instead of implementing memory handling in every agent \cite{park2023}. Because the whole conversation needs to be fed to agents at every iteration (recall that LLM APIs are stateless), this choice allows to optimize the overhead of data exchange between agents.

There are more micro-scale speech mechanisms that could be enforced in agent’s design such as contradiction, agreement and reflection. Such mechanisms are not explicitly required from agents in this first exploration, as they are organically embedded in human discourse. While such behaviors could be considered in future implementations, we believe that future model updates showcasing fidelity improvements represent a more natural integration of such behaviors into our framework.

\subsubsection{Orchestration Mechanisms}
\label{titles:orchestration}

Based on the description portraying individuals gathered in a room, peer communication is identified as the minimum required mechanism to support discussion.  In addition to peer to peer communication, which is composed of a listening phase, a sensible answer construction phase and an expressive phase, there is need to define the order in which individuals speak, and also to whom they speak, together referred to as orchestration. A trivial solution includes randomly drawing a speaker’s turn. However, early experimentation of this framework have shown inconsistency in the answer quality and discourse coherence, advising against randomly drawing speakers. For this mechanism to be realistic and relevant, there is need for implementation at the micro-scale. This turn-based communication articulates around specific agent instructions and conference room level extraction of the selected next speaker based on agent response. To this end, agent infer the next speaker from previous generated messages, using an LLM call. Content extraction is performed after every agent response, by a stateless and contextless agent entrusted to extract important features from the responses. Among these are the message recipient, content of the message, and additional properties such as the persona of additional agents that may be summoned to complement the expertise defined for current group of agents (see section \ref{titles:selfgovernanceconvergence}).

While we allow for maximum expressivity of the agents so that they can introduce new topics, revisit unresolved issues, and allow the conversation to organically adapt to emerging priorities, LLM agents have shown inconsistencies in keeping the focus on the initial topic. This limitation is partially explained by the imperfection of the data retrieval mechanism (see section  \ref{titles:selfgovernanceconvergence}), and can be addressed at the macro-scale with the implementation of continuous monitoring of responses and adjustment of prompts. Consistent with our organic approach to discourse design, refocusing and moderation is implemented at the micro-scale with the introduction of an agent to serve as the discourse moderator (or facilitator). The moderator's behavior aims to instruct the other participants to refocus on the topic if there are signs of digression, while avoiding to take part in the discussion. With this goal in mind, an analysis of the discussion is periodically conducted by the moderator agent, who summarizes the major points already addressed. Additionally, the moderator will suggest recommended discussion points that needs to be addressed.

The pace at which LLMs are updated represents a significant challenge for establishing coherent agent behavior, as their semantic expressivity increases, and their speech patterns change. Agent agreeability — the capacity for agent to systematically agree or refute statements and claims — has seen numerous changes across models and versions. Furthermore, the stance (collaborative or adversarial) of discourse represents a challenge in a framework without human intervention, in which agents self-influence their future behaviors. While a collaborative AND adversarial discourse is not impossible, our design choice presents collaboratively-generated recommendations, while taking into account positive criticism. Agents work collaboratively towards a shared objective, despite being given the possibility to refute claims from their peers, which is often a directive that agents don't follow. On the other hand, the moderator agents brings an adversarial stance, constantly evaluating the progress of discourse, and enforcing the quality and relevance of suggested courses of action.

\subsection{Self-Governance and Convergence}
\label{titles:selfgovernanceconvergence}

Having defined the discourse mechanisms at the micro- and the macro-scale, we now move on to emergent behaviors of the framework, entailing the definition of a convergence criterion. We describe the limitations of persona prompt patterns and formulate a technique for summoning new agents that allows to dynamically build an assembly, specialized for the resolution of a specific challenge. 

Persona prompt patterns (see section \ref{titles:agenticpromptengineering}) represent powerful methods for defining LLM agent’s behavioral space, and are central to the functioning of this work. The testbed for this study articulates around two example use-cases: the first portrays historical flooding events in Texas, in 2025; the second, a hypothetical extreme flood in the Midwest. Both those use-cases target to demonstrated different characteristics of the framework, yet testing the fidelity of human behavior in the face of uncertain decision-making. First explorations were conducted with handcrafted persona prompts with consideration of a fixed-sized assembly, including a mayor, a community advocate and an environmental scientist. Character design aims for specific behavior traits and restrictions pertaining to the role: for instance, the mayor is assumed as not having expert knowledge on environmental sciences, and will optimize for constraints pertaining to their standing in the assembly, including infrastructure integrity and safety of the population. As a consequence, individual assembly members optimize for a unique constraint, where the overall assembly optimizes for the union of non-redundant constraints, maximizing agent’s utility.

Despite its potential, the persona prompt technique is imperfect and prompts can show variability in the number of executed directives. This technique’s efficiency is also affected by surrounding factors such as conversation size, level of detail and ordering of the instructions. As a result, first explorations showed a large variability in the scope of the knowledge of the mayor, community advocate, and environmental scientist. This entails a significant overlap between their respective domain of expertise, and a poor added value of member interactions. Refined prompt definitions aim at more precise directives, optimizing for the enhancement of agent’s unique contribution to discourse, at the expense of a reduced scope of knowledge. As a result, the global scope of knowledge of the assembly is reduced and the minimal skill set necessary for the resolution of the challenge is not achieved, highlighting the need for more specialized agents to fill in the gaps. Handcrafting a new persona that is relevant to the decision-problem at hand, and offering a different expertise relative to the initial assembly members is then required. The anticipatory preparation of a library of agent profiles is a viable yet time-consuming solution, although LLMs offer the advantage of being able to dynamically craft a persona on demand. Moreover, the expressivity of LLMs supports a preference for LLM-issued persona prompts rather than handcrafted ones, suggesting favorable grounds for iteration-based self-governance of the framework.

This self-governed framework pattern articulates as follows: there is need for a summoning mechanism of agents that supports dynamic evolution of the agent pool towards the goal of reducing shared knowledge and increasing unique specialized knowledge. With this mechanism, the variety of the domains of expertise represented dynamically increases as new agents are summoned, while maximizing agent’s utility with more specialized persona definitions. To that avail, the summoner agent periodically reads the previously generated messages to identify explicit needs for additional agents in the conference room. Then, the summoner agent formulate a short description of the persona to summon (see Appendix \ref{titles:additionalframework}). The framework generates a persona prompt from the agent-given description, and creates another agent entity that is registered in the network and introduced in the conference room. This self-governance mechanism provides a new performance and evaluation metric, defined from an aggregate of framework variables, such as the size of the assembly, the different agent roles and the participation rate of every agent. This contrasts with a more straightforward convergence criterion, built from the quality, relevance and breadth of recommendations. Hence, convergence can include stability of the assembly distribution, a defined set of domain expertise represented, or a minimal assembly that satisfies a relevant assessment of the decision context while maximizing unique agent utility.

\subsection{Explorative Discourse}
\label{titles:decision}

We mention in section \ref{titles:selfgovernanceconvergence} that the quality, relevance, and breadth of the framework-made recommendations is of importance for performance evaluation. In addition, the stability and consistency of generated actionable measures can also constitute a convergence criterion. However, complications in decision-making arise from the unexpectedness exerted by uncertainty. Taleb \cite{taleb2007} refers to unpredictability as “outside the realm of expectation,” which also carries a semantically strong attribute of unexpectedness. Unexpectedness is anchored in human perception owing to the availability bias, defined by behavioral sciences as an imprecise evaluation of risks and event likelihood based on the scope of available information \cite{lieder2018}. In other words, humans imagine and conceive risks and events that they experience or are knowledgeable about, at the involuntary expense of unlikelier ones. For instance, well established examples from behavioral sciences include the misconception that suicides are less frequent than homicides as a result of a lower media representation \cite{efendic2021}.

With consideration to the changing patterns in the occurrence and strength of extreme natural hazards, the absence of historical records crystallizes an availability bias to which human beings are subject. This causes humans to sometimes misevaluate the risk of extreme events happening, resulting in hindered preparedness to such events. Generative AI then represents a compelling method to evaluate alternate pathways, as it does not suffer from the same limitations than randomized controlled trials (resource cost, human bandwidth, ethical considerations). Indeed, recent work shows that LLMs show characteristics of world models \cite{yang2024, hao2023}, entailing a favorable environment for agents to perform experiments as they’re provided with simulated sensors and actuators. In addition to the horizontal scalability offered by multi-agent architectures, this makes our framework compelling for the exploration of a wide breadth of actionable pathways in face of a complex decision-making problem. Traditional methods rely on physical modeling whose performance is hampered by the nonstationarity of modeled processes, whereas agentic LLMs continuously adapt through in-context learning and counterfactual evaluation through variations in the input scenario. Hence, agentic LLM offers novel perspectives based on breadth-first exploration of alternatives mirroring real-world decision making.

LLMs exhibit great expressivity through natural language, that offer favorable grounds for defining task optimization constraints with the use of persona prompts. Indeed, the personal and professional objectives, values, and aspirations of generated persona organically shape optimization constraints, fostering a closer representation of real-world complex decision needs. These include the practicality and resource limitations of implementing strategies, the capacity for generated strategies to improve ecological and engineering resilience (formally defined using dynamical systems theory \cite{srinivasan2015}), and the inclusion of social and economic justice in the distribution of benefits and burdens resulting from generated strategies. Specifically, the community advocate LLM agent solely aims to organically address the underrepresentation of social justice and equity in the face of extreme events mitigation \cite{marino2023}. On the other hand, the agent impersonating the mayor must think about the structural and economical integrity of buildings and facilities, vital to the rapid recovery of the system after a shock. In this regard, our framework mirrors real-world decision-making where outcomes must balance technical feasibility with broader societal impacts, while accounting for uneven distribution of resources and burdens.

\section{Application}
\label{titles:methods}
In this section, we develop two use cases to demonstrate the applicability of this framework, as well as its sensitivity to uncertainty, policy, and risk posture. These two use cases allow to draw qualitative conclusions about the operation of a discourse agentic LLM framework in uncertain complex scenario, despite the lack for well-documented evaluation metrics. Such challenges are further discussed in section \ref{titles:relevance}. First, the agents in the starting assembly are presented. Second, we describe the application use-case, that uses a recollection of events from the floods in Texas in 2025. Last, a hypothetical flood scenario in the Midwest is used to explore the response changes for varying uncertainty conditions.

\subsection{Initial Pool of Agent Persona}
\subsubsection{Local Government Representation}
\label{titles:mayor}

The first agent personality that was designed in this framework is the mayor of the virtual township. The mayor represents a stakeholder or decision maker who is trusted with making the final recommendations. The mayor is instructed to rely on personal knowledge, that does not include any particular environmental science priors. Additionally, this agent’s design includes a directive to consider consequences of the recommendations made. Last, the possibility for this agent to disagree with its peers and challenge their points is specified, primarily as an initiative to steer the LLM away from its agreeing and compliant baseline behavior, which is restrictive in this use case.

\subsubsection{Environmental Science Representation}
\label{titles:scientist}
This agent is described as an urban and environmental engineer, designed to bring a scientific aspect to the discourse and provide sources and knowledgeable information. Therefore, there is future work opportunities of using such a persona to integrate advanced features such as fact verification through content extraction from research material, leveraging methods such as Retrieval Augmented Generation (RAG). Similarly to the first role presented, this agent is also encouraged to disagree with its peers and challenge their arguments, if needed. Last, we seek to improve this agent’s reasoning capacity to reinforce its relevance as a science representative, using prompt engineering. Chain of Thoughts (CoT) is a prompt pattern that enforces a LLM to produce intermediate \textit{thoughts} before producing the final output. CoT is a well established technique that has shown improvements to the reasoning quality and global precision of the model \cite{wei2023}. Using CoT, we enhance the scientific relevance of the scientist's persona.

\subsubsection{Community Advocacy Representation}
\label{spokesperson}

In contrast to the most common optimization constraints presented, social equity and justice in the face of extreme natural hazards is not integrated in the input scenario. The underrepresentation of social equity and justice in real-world discussions and decision outcomes is well documented, particularly in the face of extreme events challenges \cite{ipcc2023, marino2023}. Following that perspective and on the grounds of mirroring real-world situations, we design a single initial agent that optimizes for social equity and justice.  Hence, this third agent is designed to incorporate the role of an elected spokesperson that represents the low-income neighborhoods and at-risk communities in the township. Similar to the mayor, this spokesperson does not have particular environmental science knowledge, and is offered the possibility to disagree with its peers. However, this agent is designed to raise concerns about the safety and consideration of the communities it represents, and in appropriate situations it challenges its peers on points or recommendations.

This choice of implementation entails multiple remarks: the fact that such mitigation aspects are embedded in a single initial agent does not restrict other summoned agents to optimize for the same objective. In the initial pool, the agent that supports social equity and justice as a primary requirement carries the same importance as other optimization constraints, preventing it from being a secondary objective. There are other relevant implementation choices that are discussed in  section \ref{titles:discussion}.

\subsection{Application Use-Case: Texas Floods of 2025}
\label{titles:secondscenario}
This use case aims to demonstrate the applicability of the framework on a historic use case. The recollection of events portrayed in the input scenario below were gathered over multiple sources \cite{ap_texas_flash_flood_risk, ap_texas_floods_timeline, npr_texas_floods_timeline, nasa_earthdata_tx_flood}. The framework leverages GPT-4.1-mini, GPT-5-mini and GPT-5 for agents backend. The training data cutoff for those models is the Fall of 2024 for the most recent. Thus, the events that occurred in July of 2025 in Texas are absent from those model's knowledge, protecting against any form of data contamination.

\subsubsection{Research Context}
\label{titles:researchcontexttf}
This scenario collects events that happened during the flooding events of Kerrville, TX, in July 2025. The timestamp cutoff — that represents the time after which real-world actions are not included in the input scenario — is 2025 July 4th at 6 a.m. This timestamp corresponds to the first reported response from decision-makers, allowing the agentic framework to generate an action plan that can be compared against the real-world events that followed.

This scenario is constructed to evaluate whether the framework offers recommendations that are consistent with an observed response. To that avail, the evaluation criterion we adopt for this use case could be defined as qualitative comparability. In essence, the output is qualified as comparable if it includes all the decisions that were made in the real-world events, that are presented in Appendix \ref{titles:evaluation_baseline}. While we acknowledge the biases and limitations of such a method, a point-by-point comparison would be impractical, tedious and imprecise (due to: synonyms; terminologies; semantic, lexical and conceptual overlaps). Moreover, agentic LLM evaluation methods are still strongly underrepresented in the current state of the art. To our knowledge, there is no well-documented and proven evaluation method for challenges such as the one portrayed in this study. We discuss the implications and opportunities of such limitations in section \ref{titles:coherence}.

In this context, the framework is run fifteen times using the input scenario described in section \ref{titles:texasfloodsscenario}. Resulting summaries are condensed into a final assessment by a single GPT-5 agent. Evaluation is performed by a single human operator, thus preserving vocabulary and intent. This final assessment is reported in Appendix \ref{titles:summary_action}, and compared against the baseline presented in Appendix \ref{titles:evaluation_baseline}.

\subsubsection{Scenario}
\label{titles:texasfloodsscenario}
The scenario is presented in the exact same form as the one fed to our framework. The National Weather Service (NWS), National Oceanic and Atmospheric Administration (NOAA), and the United States Geological Survey (USGS) are agencies of the United States of America.

\textit{In Kerrville, TX: July 3rd at 1:45 pm, a flood watch is issued by NWS. During the day, the Texas Division of Emergency Management escalates the State Operations Center to level 2 and mobilizes additional state emergency resources during the afternoon. July 4th, at 1:15 am, NWS escalates the warning to a considerable flash flood warning triggering wireless emergency alerts and NOAA weather radio alarms. CodeRed is requested at 4:22 am, and sent at 5:30 am. During the night, the flood warning was escalated to very dangerous, citing 3 to 7 inches of rain in 2-3 hours. This is July 4th 2025, 6 a.m. Determine a contingency plan, starting with listing potential prior factors for disasters, including economic factors, human factors, social factors, etc. Identify immediate risks, and conduct a thorough analysis of a course of actions. Here is a USGS reading of the Guadalupe river,in feet, at the Kerrville gauge: $<$USGS file content$>$}

\subsection{Sensitivity Use-Case: Hypothetical Midwest Floods}
\label{titles:scenario}

\subsubsection{Research Context}
\label{titles:researchcontexthm}
This second scenario explores the sensitivity of the framework to varying uncertainty, revolving around low-probability and high-risk weather events. With the first use-case focusing on real-world applicability, this second use case sets in a hypothetical setting. This allows to further explore the response of the framework to missing or incomplete information. The following text describes the hypothetical extreme rain event situation, the context in which it occurs, and serves as a bootstrap prompt for LLM agent interactions. This scenario was built and inspired following some real-world challenges highlighted by foundational climate assessment reports \cite{mach2023, burkett2014, jones2014, ipcc2023, marino2023, clarke2018} and translated to optimization constraints in the presented scenario. Examples include preservation of vital services (energy, drinking water), operation of commercial routes and safety of the population. 
For this scenario, 15 executions of the framework were conducted, 5 for each parameter of the probability of the event happening: 50\%, 75\% and 90\%.

\subsubsection{Scenario}
\textit{We are in a US Midwestern township of half a million inhabitants. A large river flows through this township which is also fed by a large watershed in which the township is located. There is forecast for very heavy rain and possibility of flooding at large scale. Consider that the probability of flooding is \textless probability parameter\textgreater. The township needs to make anticipatory decisions to respond to the impending event to minimize the impact of floods but also keep in mind the needs of the community which relies on the river water. A reservoir downstream of the town supports potable water needs, provides energy, and recreational needs. The river also supports navigational and commercial traffic. Decisions must address the management of reservoir levels. At the end of the conversation, a strategy needs to be developed to manage the outcome of the flood keeping in mind uncertainty of the event, noting that different probabilities of risks may justify different approaches, and the associated advantages and drawbacks of the decision variables.}

\section{Results}
\label{titles:results}

\subsection{Texas Floods of 2025}

\subsubsection{Observations}
\label{titles:interpretationtf}
The condensed assessment generated by the framework across 15 runs is comprehensive, consistent with real-world actions, and integrate course of actions that are absent from real-world events. The resulting action plan presents four main phases of recommended actions: first 3 hours, 3-6 hours, 6-24 hours, 24-72 hours. The first phase strongly emphasizes hydrologic confirmation, as well as the constitution of supervising Unified Command with specific sub-divisions. In essence, results show the first 3 hours to be instrumental in establishing a strong and organized response, focused towards starting evacuation, opening shelters and protecting critical infrastructures. The second phase introduces request for state and federal aid, and emphasizes continued evacuation, and search and rescue efforts. Additionally, organization is refined with measures such as triage points for emergency services, evacuation monitoring, and confirmation of shelter operations. The third phase predicts the start of recovery efforts, shifting from life-preserving directives to medical and psychological assistance. Critical organization directives are still at the center of the plan, such as securing food and other crucial supplies, assessment of at-risk infrastructures, and quality checks of drinking water contamination. Last, the final phase continues damage assessments to infrastructures, and proposes the establishment of disaster assistance centers. Moreover, the emphasis is brought over a phased re-entry of cleared areas residents, as well as the transition from shelters to temporary housing.

These results represent a comprehensive approach, whose steps are comparable to an Incident Command System approach \cite{jensen2016}. The emphasis is placed on hydrologic confirmation (with recommendations such as "confirm hydrologic reality by validating gauge readings and classifying the event
as a catastrophic flash flood"), and recommendations are informed from data. The projected operational response plan is consistent with real-world reported actions, presented in Appendix \ref{titles:evaluation_baseline}. With respect to the challenge of accurately collecting all historical decisions made, our results include plausible course of actions we could not find in real-world reported events.

\subsubsection{Observations on Agent Behavior}
\label{titles:interpretations_agent_tf}
Variability is a central element of agentic LLM discourse. Ensuring balanced exploration while maintaining comprehensive decision-making is challenging, especially when different framework executions can result in different outcome trajectories. We observe a trade-off between the varied exploration offered in individual executions and the condensed assessment. Individual executions benefit from a varying degree of exploration for specific subtasks, as a result of the stochastic natures of LLMs. Distinct features are unique to each run, and do not necessarily appear in condensed assessments, whose objective is to crystallize recurring recommendations to ensure coherence. In individual runs, specific summoned agent combinations offer deeper perspectives on specific aspects of problem-solving. The following list of observations is not exhaustive, but aims to demonstrates how agentic combinations create new synergies within simulated human decision-making.

The combination of an emergency planner agent and the scientist agent yields highly structured planning. The combination of a hydrologist agent and the scientist agent brings a major focus over data-informed decisions, and the requirement for additional data feeds. The combination of an emergency planner agent and a GIS agent results in routes tagged as at-risk in the context of major projected floods, and were recommended closed. We verified the proposed route closure, and confirm that these axes were in flood-prone areas. The combination of the community advocate agent and the mayor agent yields equity-focused evacuation and takes into account socioeconomic factors in flood mapping. Moreover, this agent combination proposes the definition of evacuation routes, and an estimation of necessary evacuation.

\subsection{Hypothetical Midwest Floods}
First, some remarks need outlining. As a result of missing or incomplete information in the input scenario (e.g. precise geographical setting, forecast rain quantity, details on vital infrastructures and activities), a strongly recurring pattern drives agents to request additional data. Owing to the intentional opacity of the scenario, agents present fabricated field data, such as reservoir level estimates and stage measurements. While these fabrications could be seen as harmful in real-world settings, we advance that in this setting they reflect concerns for a minimum amount of data required for a correct situation assessment. The exercise of providing sensible action plan recommendations without proper data, and with vague information, is near impossible. We justify the tolerance for data fabrication in this case, on the grounds of observing the framework's response to strong uncertainty and dire constraints. Additionally, such fabrications were not observed during executions of the first use-case (2025 Texas Floods), grounded in a real-world historical setting.

\subsubsection{Interpretations}
For the sake of convenience, results for the scenario in which the probability parameter is 50\%, 75\%, and 90\%, will be referred to as S50, S75 and S90, respectively. Results for S75, S90 and S50 are presented over the next paragraphs, in this order.

\begin{figure}[!h]
\centering
\includegraphics[width=1\linewidth]{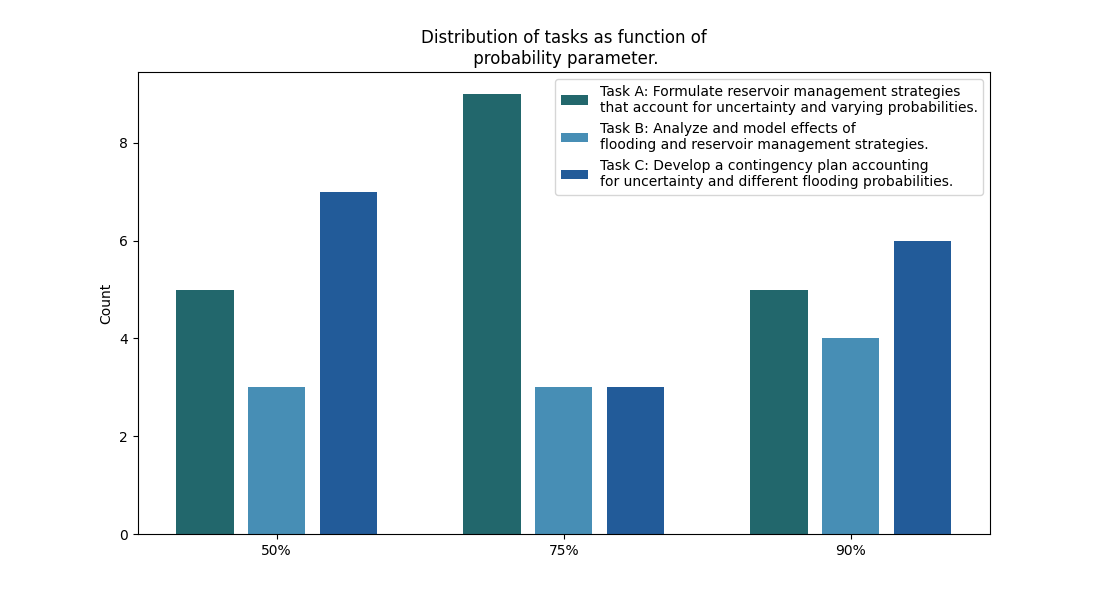}
\caption{Distribution of selected tasks as a function of the probability parameter. The horizontal axis represents the probability parameter in the input scenario, respectively 50\%, 75\% and 90\%. Additional supporting data can be found in Appendix \ref{titles:additionaltable} (Table \ref{tab:simulation_results}).}
\label{fig:task_probability}
\end{figure}

First, the conversations and summaries generated for \textbf{S75} demonstrate a defensive posture towards the flooding risk. This posture is illustrated by: the recommendation of proactive drawdowns aiming to create a predefined buffer in reservoir storage; clear definition of triggers for escalating alert levels, evacuation, increase of reservoir releases, illustrated by Fig. \ref{fig:task_probability}. Evacuation or pre-evacuation phases are part of the action plan in 4 out of 5 runs, despite the presence of a dedicated agent for that task. In this setting, reducing the risk of flood is paramount, even if it needs to be done at the expense of hydropower generation, navigation, and recreational activities. The key takeaways from S75 is that the risk is real: dam safety and downstream protection take precedence over revenue. In this scenario, the cost of being wrong about the flood is higher than the cost of losing money over electricity and commerce. Therefore, the recommended action plan is articulated around reservoir operations mainly, while evacuation is still strongly represented.

For the \textbf{S90} scenario, the stance evolves from defensive to aggressive response. A probability of 90\% signifies that the threat is real and must be received as such. As a result, agents discuss drastic measures such as reservoir release up to the maximum legal limit, and discuss that using the spillway above those limits requires a one hour warning, as well as legal authorizations. In S90, balancing transparency and panic control is paramount, illustrated by the definition of multiple stages of urgency in emergency communication. Fig. \ref{fig:task_probability} shows that for S90, tasks are almost evenly distributed, harmonizing with the agent's comprehensive approach.

Last, the 50\% probability scenario represents the most interesting setting. While \textbf{S50} represents the lowest risk of all three tested scenarios, it is the most uncertain. In this scenario, agents see themselves under additional pressure resulting from the increase in uncertainty. Disaster Risk Reduction science gives the following definition of risk: "the potential loss of life, injury, or destroyed or damaged assets which could occur to a system, society or a community in a specific period of time, determined probabilistically as a function of hazard, exposure, vulnerability and capacity" \cite{disasterrisk}. In S50, the uncertainty lies in the probability of the event happening, which strongly increases Risk as overreactions or errors from decision-makers could result in grave consequences. For instance, if drastic preemptive measures such as activating the dam's spillways were followed by the storm not hitting the town, severe financial, legal, and human constraints would entail: partial loss of economical activity, drinking water unavailability, potential loss of hydroelectric power. In response to those constraints, the agents prioritize a cautious approach with adaptive measures including the definition of multi-stage triggers. All runs report action plans including light to moderate preemptive releases of the reservoir, with a strict minimum level that must be conserved. In contrast, Fig. \ref{fig:task_probability} shows that contingency measures are more represented than other tasks, and is illustrated by 3 conversations out of 5 that define evacuation and shelter preparedness as central measures, coordinated by a summoned emergency management agent. There is a representational difference between the distribution of tasks, and the recommended actions brought forward by agents, which indicates a difference in perspective between the task-breaker agent and agents engaged in discourse. The task-breaker agent has a generic prompt whose instructions are shared by most current LLM frameworks, while persona agents are crafted towards the goal of engaging in human-like decision making. We advance that the perception of urgency and risk manifested by discourse agents is an emergent behavior of agentic discourse, supporting the hypotheses we formulate in section \ref{titles:introcomplexity}.

Overall, these 15 executions with varying probabilities of the event happening introduce what is decision-making under uncertainty: making decisions when the data is not perfect. This is grounded in agents requiring more data in almost every run. Because the agents are presented with a scenario that carries a significant risk, they fabricate data in order to carry on with their assessment. Conceiving potential solutions without data is a challenging task that entails a significant number of hypothetical branches to explore. Nevertheless, the discussions generated by agents offer an interesting perspective: the safety of this hypothetical town doesn't depend on the dam itself, but from its operation. There is a line that separates preserving potential revenue and saving lives, and agents clearly explore the competing interests of the major actors in this scenario. A concluding remark stem from agents worrying about paperwork and security permissions. Activating a dam's spillway above its restricted limit requires emergency authorizations as well as one hour of alert broadcasts. These constraints can hardly be integrated in numerical physics-based models, nevertheless is included as a central element in the recommended action plans. This encapsulates that even in simulations, bureaucracy is a constraint, which we present as a qualitative contribution that agentic LLMs capture human decision-making, to some degree.

\begin{figure}[!h]
\centering
\includegraphics[width=\linewidth]{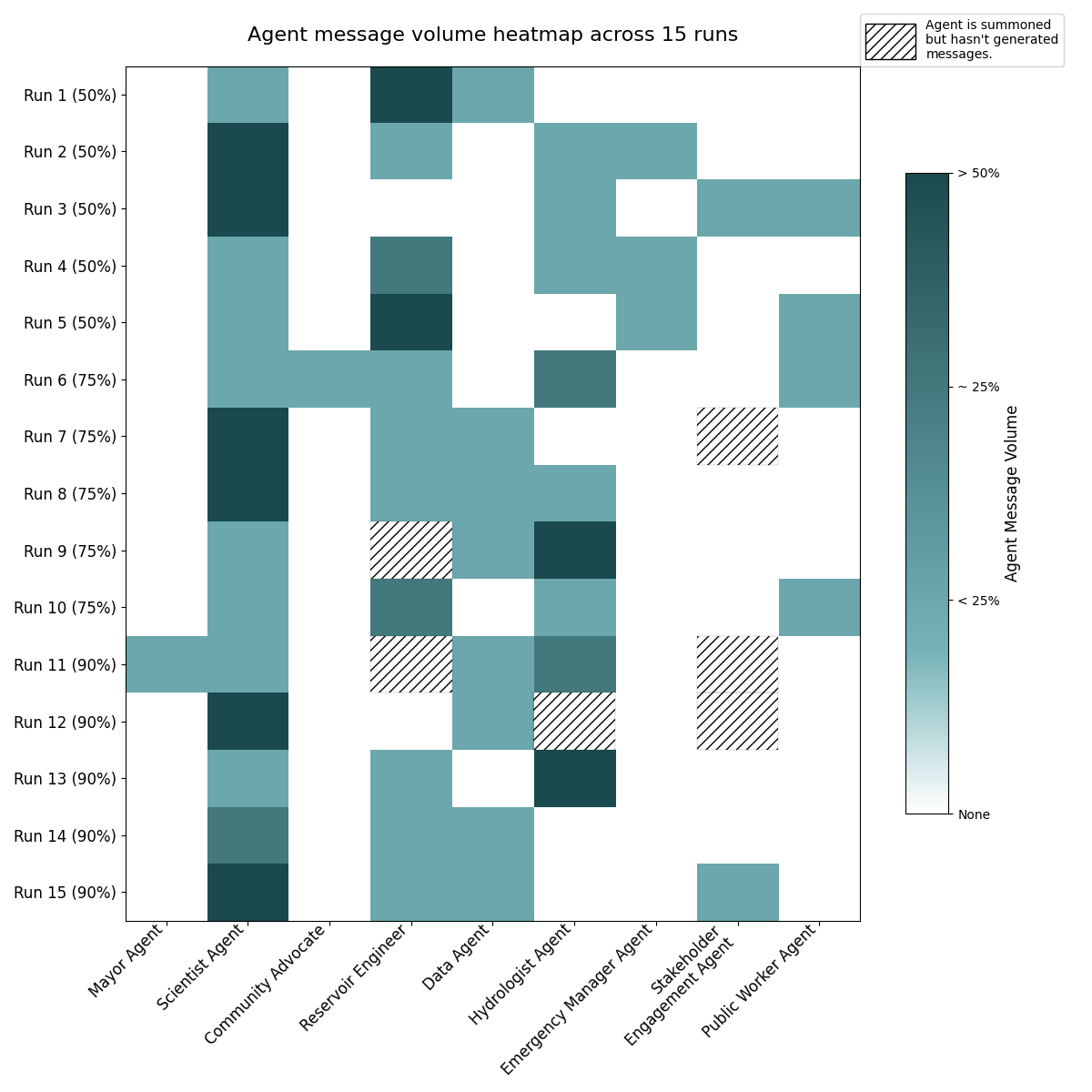}
\caption{Heatmap of the volume of messages generated by agents (on the x axis), for each run, for every probability parameter in the input (on the y axis). The percentage cutoff is indicated on the colorbar, where \textit{None} indicates the agent hasn't interacted for the run. The mayor agent, scientist agent, and community advocate agent are present in every run; every other agent is summoned. Despite being summoned, an agent might not generate messages, if never chosen by the orchestrator: this case is shown on the heatmap as a hatched cell.}
\label{fig:agent_heatmap_midwest}
\end{figure}

\subsubsection{Observations on Agent Behavior}
\label{titles:agent_interp}
The analysis of agentic conversations yield promising conclusions, as we observed in the previous section. Moreover, we advance that observations about the distribution of summoned and active agents can also yield interesting remarks. Indeed, we explain in section \ref{titles:introcomplexity} that our choices for a multi-agent architecture are motivated by the multi-objective nature of challenges such as disaster resilience. In order to support those observations, Fig. \ref{fig:agent_heatmap_midwest} presents the volume of messages generated by each agent, for each run. If a cell in the heatmap shows \textit{None}, the agent hasn't had a chance to participate, or was not summoned in the corresponding run. In the previous section, we mentioned that 3 out of 5 S50 runs include evacuation in the action plan, more specifically runs 2, 4 and 5. Now, we see in Fig. \ref{fig:agent_heatmap_midwest} that an emergency manager agent was summoned in each of these runs, which allows us to advance that this agent is positively contributing to the addition of evacuation to the action plan. This is confirmed by a message generated by the summoner agent — the agent responsible for identifying underrepresented skill profiles : \textit{"I have summoned emergency\_response\_agent to draft evacuation thresholds, communication protocols, potable water backups, stakeholder coordination, and monitoring cadence. Emergency\_response\_agent should provide these so the scientist\_agent can integrate them into contingency plans and finalize the flood management strategy."} In contrast, in S75 evacuation is mentioned in every run and included in the action plan 4 out of 5 runs. Nevertheless, emergency manager agents are never summoned in these runs, indicating that the urgency of the situation caused other agents to address this challenge first.

Another example of the importance of agents is shows in the run 1 of S75 (labeled as Run 6 on Fig. \ref{fig:agent_heatmap_midwest}). To our knowledge, this run is the only one that includes priority evacuation for low-income neighborhood, and shows activity of the community advocate agent, which states: "\textit{Recreational and navigational concerns are secondary during a high flood-risk event. We favor moderate risk tolerance favoring proactive releases to reduce flood damage in low-income neighborhoods prone to flooding.}" While social equity and justice has gained representation over the past decade, there still are challenges to be addressed \cite{malini2022}. We advance that the community advocate agent brings a novel perspective to the inclusion of social equity and justice in the face of decision-making in the face of natural hazards response. More broadly, the agent's persona are built from realistic actors of decision-making in order to include multiple competing priorities at the core of the scenario: reservoir manager agents require quick decisions in order to implement actions; emergency management agents have no concern for economical activities, and strive to ensure population safety and often require timeline estimations; hydrologists need more data, run models and crunch numbers in order to provide the best scientific assessment of the situation. Each agent embodies specific roles withing decision-making, anchoring agent personas as a crucial part of this study. Therefore, we acknowledge that the constitution of the initial pool of agents, the distribution of summoned agents in function of the nature of the scenario, and the latitude given to the summoner agent are major parameters to agentic discourse frameworks, that would benefit from deepened attention in the future.

\section{Stress Points: Understanding and Addressing LLM Vulnerabilities}
\label{titles:modesfailure}

Since their recent popularization, LLMs have been under close scrutiny and have shown many conceptual and operational flaws, addressed by model updates, better training data and structural concepts aiming to mitigate some of these flaws. In this section, we detail the most common modes of failure that can impact the multi-agent discourse outcomes.

\subsection{Hallucination}
\label{titles:hallucinations}
When LLMs fabricate facts, premises and claims, they are known to hallucinate. These erroneous claims can also be defined as false positives or negatives, hinting at a wrong sense of confidence exerted by LLMs. What causes hallucinations in LLMs is still at the center of multiple research studies, although recent work has shown that they are believed to stem from deficiencies in the data retrieval mechanisms, also referred to as prompting \cite{feng2025rethinking}. In subsequent sections, we build upon those findings to theorize why agentic LLM show reduced hallucinations. Fine-tuning is a well-documented method to mitigate hallucination, sacrificing factuality in exchange for adhering to human-centered preferences \cite{plaat2025}. Yet, there are findings that formally prove that hallucination-free LLMs can’t exist \cite{banerjee2024, xu2025}. In accordance with the human-like desired behaviors of LLMs, we list here mitigation strategies that are deeply connected with human inspiration at the core. Self-confidence evaluation has shown that LLMs have an unbalanced assessment of their own generations \cite{pawitan2025, fadeeva2023, fu2025}, leading to the conclusion that LLMs are far more efficient and accurate at verifying facts and evidence rather than generating accurate answers \cite{huang2023, huang2024}. Confidence and self-confidence evaluation methods articulate around the generation of confidence score for LLM-generated answers, as well as qualitative assessment such as continuity and coherence in answers as seen in \cite{pawitan2025}. Self-confidence can be measured using various confidence quantification metrics, such as next-token and next-sequence prediction distribution, ensemble based methods and density based methods \cite{fadeeva2023}. In contrast, there are methods focusing on quantifying the uncertainty of generated answers, such as conformal prediction \cite{ye2024, liang2025}. Results are heterogeneous and harmonize to conclude that LLMs have a limited ability to self-assess their own generated responses, hinting that external sources to check against are more reliable.

Retrieval-augmented generation (RAG) leverages external data sources (articles, databases, data repositories) in the inference pipeline in order to provide fact-checked premises to enhance LLMs' reasoning process. Retrieved data is converted to vector embeddings and stored for later use during inference. Fusion of the generated data can be performed at multiple stages during inference, referred to as query-based fusion, logits-based fusion or latent fusion \cite{wu2025}. RAG models consistently show performance improvements over baseline models and reduced hallucination \cite{mala2025, ayala2024, shuster2021}.

Finetuning is a well documented approach for reducing LLM hallucination and improve model performance. However, finetuning presents multiple shortcomings that stem from the No Free Lunch (NFL) theorem \cite{wolpert1997}, and are rarely mentioned. Indeed, the NFL theorem proves that no learning algorithm is universally better than any other when averaged over all possible problems. While there is no formal proof of the Neural NFL theorem, core implications remain and suggest that finetuning imposes a tradeoff between generalizability and task performance \cite{goldblum2024, chollet2019}. Because of the tedious efforts necessary to train models, the non-transferability to newly published models, and the permanent specialization of finetuned models, we do not envision finetuning as a relevant method to address hallucination in this work. In contrast, we argue that temporary specialization via persona agents is a far more effective strategy to counter hallucinations, while preserving flexibility in reasoning performance and domain of specialization.

Recent work advances that “prompt learning methods do not change the parameters of the model; in order to use the data that is generated by inference-time approaches, finetuning must be used.” \cite{plaat2025} While we agree with such a statement, we’d like to bring some clarity to our conceptual view of hallucinations. A trained LLM can be viewed as a matrix that represents a contextually-aware next-token prediction function. Thus, long-context inference can be reduced to conditioned sampling, where the conditioning variables integrate previous tokens in the context window. With this perspective, hallucinations are represented by the inability of a LLM to correctly infer the next-token given a certain context. In other words, hallucinations can be described as the inability to access certain parts of the next-token distribution, given the context. This view is aligned with recent work \cite{feng2025rethinking}, that "systematically demonstrate that the model failures [...] arise from the suboptimal retrieval of parametric knowledge in LLMs, that is, their inconsistent ability to reliably access knowledge encoded in their parameters." While prompt engineering does not change the parameters in the models, it provides another contextual \textit{pathway} in the successive conditioning steps, hence allowing to reach other parts of the distribution. Because sub-parts of the distribution could be assimilated to different learnt behaviors, prompt engineering entails strong behavioral modifications without the need to finetune. When persona prompt patterns are applied to multiple agents, they are used as non-permanently specialized LLMs, and mitigate constraints imposed by the NFL theorem, while maximizing the utility of individual agents.

\subsection{Coherence}

\label{titles:coherence}

Previous paragraphs of this section address the structural and semantic challenges posed by monolithic and agentic LLMs. In this section, we present a characterization of coherence, and its impact on the quality of discourse, and consistency of the generated responses.

\subsubsection{Moderator Agent}

On the grounds of the organic construction of the framework, consistency is enforced by the moderator agent, at the core of the iteration loop. Specifically, this agents aims to prevent goal drift, which is defined as an increasing deviation from the initial topic of interest. Both repetitions and goal drifts constitute a decrease in consistency as discourse unfolds, especially in no-human-in-the-loop autoregressive workflows. This characteristic represents a major challenge of agentic discourse frameworks in itself, for which the moderator agent is a countermeasure. This agent periodically runs an assessment of the conversation, extracting explored recommendations, listing explored tasks, and reviewing messages that do not positively contribute to discourse. The moderator’s assessment contained a compressed analysis of the current state of discourse, guidelines for unexplored strategies, and recommended topics of discussions.

\subsubsection{Testing Framework}

Unit and integration testing is another method to ensure coherence, that is well-documented in Computer Science. Ensuring the coherence of the whole framework is first achieved by ensuring the coherence of its components. Adopting the standpoint of software engineering, conformity and regression tests represent an interesting perspective to ensure the consistency of individual agents. Whereas LLMs are thoroughly compared in benchmarks, the current testing procedures do not focus on the stability of expected behaviors, but rather on predictive performance. In this work, consistency, coherence and believability of persona characters are key to support sensible decision support. Existing methods include agent questionnaires that can be leveraged to rank the believability of agents \cite{park2023}, although posing a challenge of human annotators bandwidth and scalability. On the other hand, BIG-Bench \cite{bbh} includes a testbed of various tasks worth exploring, despite omitting agentic workflows. Additionally, most major test benchmarks such as BIG-Bench focus on reasoning and logical tasks, rather than discourse capabilities. Recent work has focused on monitoring agent personas in real-time during discourse, and has demonstrated that maintaining their consistency is challenging an represents a relevant issue \cite{bhandari2025can}. These findings support the need for discourse-focused agentic benchmarks, harmonizing with the recent growing interest in multi-agent LLM implementations.

\subsection{Relevance}

\label{titles:relevance}

\subsubsection{LLM Evaluation}

LLMs belong to the Natural Language Processing class of Deep Learning algorithms, although their growing popularity and ubiquity suggest they represent a class of their own. Nevertheless, in contrast with nearly all historic models, LLMs are text processors. Whereas classical models are built in a numerical frame of reference, LLMs are built upon token sequence projections, that encode natural language into a high-dimension vector space. This means that LLMs also have an underlying numerical layer — the one that encode the latent representation embeddings and the model parameters — albeit non-explainable, high-dimensional, and still considered a black box. As a result, there is no reproducible and statistically significant method to associate text sequence outputs with LLM embeddings and/or model parameters, strongly reducing opportunities for classical model evaluation. Whereas traditional mathematical or physics-based models can be described with input variables, output variables and parameters, LLMs can’t benefit from such a formalism. As a result, conventional evaluation metrics such as $R^2$ score, median average error and root mean squared error don’t apply in this work.

\subsubsection{Emerging Methods}

In this context of non-numerical model assessments, in which outputs are potentially lengthy and model parameters are unexplainable, outside-the-box thinking is necessary. Recent work has presented methods that mainly articulate around: log-probability distributions of token or token-sequence as a measure of uncertainty \cite{fadeeva2023, fu2025}; self-evaluation through (i) continued response consistency \cite{pawitan2025} (ii) confidence score evaluation \cite{liu2024mind, yang2024} and (iii) information-theory metrics \cite{chang2024evince, chang2024socra, riedl2025}. However, such methods are still criticized, particularly for the bias that such methods pose \cite{xuwenda2024}. Nevertheless, LLMs are the state of the art of NLP to this date \cite{mizrahi2024}, advocating there is no better than LLMs to evaluate text.

\subsection{Adversarial Prompt Injection} \label{titles:adversarial}
Adversarial prompts generally refer to attempts of modifying the overall behavior of LLM agents beyond their own guardrails through prompts, and is also referred to as prompt injection \cite{liu2025prompt}. Well-known examples include the generation of unethical responses by providing an emotional context \cite{gandhi2025, sahler2024}. Adversarial prompts are less effective on newer models as a result of an increased reasoning capacity which provides better countermeasures for LLM to evaluate purposes, and increased semantic security to prompt injection \cite{sang2024}. Other  detection methods and countermeasures are developed to adaptively address the growing ubiquity of agentic LLM in modern applications \cite{debenedetti2025, rehberger2024, gosmar2025}. While we don’t neglect nor belittle such a limitation, it is reasonable to consider it less relevant in our case, our framework being open-source and in development.

\subsection{Biases} \label{titles:biases}

For the sake of simplicity and reproducibility, the framework presented in this work leverages off-the-shelf versions of GPT-4.1-mini, GPT-5-mini and GPT-5. Whilst OpenAI’s GPT models dominate the technical benchmarks since the advent of LLMs, those models still showcase technical limitations as discussed above. In the context of this work, scientific relevance is crucial which can be severely hampered by such limitations: notably, the availability bias (defined in \ref{titles:decision}). Because LLMs are trained with human-generated content, they could be subject to a form of availability bias. Reinforced by the better ability of LLMs to verify facts and evidence compared with their ability to generate accurate content \cite{huang2023, huang2024}, a fact-checking component based on trusted and reliable sources is indicated to assess such a challenge. A fact checker agent that browses and parses through research material in order to validate or refute claims made by the agents is an example of such a component. There are many repositories that could be leverage for such a task, such as arXiv and Researchgate. While modern LLM chatbots present such capabilities and can provide academic references to support their arguments, API-level LLMs do not, as is discussed in sections \ref{titles:agentdesign}.

\subsection{Overestimation of Capabilities}
Due to the quality of LLM text production, readers aren’t immune from validation bias and could underestimate potential LLM errors, and involuntarily lower their critical thinking. While this issue creates the requirement for safeguards in operational settings, we consider it less relevant at the current stage of the framework, aimed at establishing feasibility.

\section{Discussion}
\label{titles:discussion}

We have developed and presented a concept for a self-governed multi-agent LLM framework that simulates complex decision discourse under uncertainty by impersonating human-like stakeholders. Two application use cases are presented, for which results are discussed, showing promise for human-like decision-making. While the previous section presented known mode of failures of LLMs, in this section we discuss potential development that would follow the inevitable popularization of agentic LLM frameworks. To that avail, we present further discussions articulating around the framework design, major areas of LLM research, takeaways from the results, perspectives on future developments, and possible implications for other areas of research.

\subsection{A Dynamical System Perspective}
\label{titles:dynamic}

Resilience-focused decision-making is often described conceptually, absent of mathematical formalism and clear foundational decision framework. Following this approach, decision-making is described as a succession of actions that either positively or negatively contributes to resilience, where resilience is defined as the maneuverable space constrained between biophysical and social stressors \cite{burkett2014}. Despite being defined using concepts of decision science, dynamical systems analysis offers a categorization of resilience with domain of attraction regimes and stability analysis of critical points \cite{srinivasan2015}. In Dynamical Systems Theory (DST) terms, the different states of resilience can be represented as critical points of the systems, in which attraction regimes is conditional of decisions. In this context, resilience is achieved by causal interventions  — either setting path on another trajectory, or  by creating a bifurcation in the system that shifts attraction regimes \cite{srinivasan2015}. DST is tightly connected to causality, requiring controlled experiments in order to test causality assumptions. Well documented limitations of such methods include the necessary resources to conduct controlled experiments, especially in CHANS' modeling \cite{kimmel2021, haag2001}. In this context, we advance that agentic LLM environments represent a sandbox for human-like decision-making, in which testing interventions and conducting controlled experiments costs significantly less resources. While we can't argue that agentic LLM don't presently offer the same level of detail and trustworthiness, we advocate that their data generation capacity represents interesting developments for DST. We conducted simple trials aiming to: probe our framework's ability to understand dynamical system concepts; characterize an input scenario using DST formalism; propose stability projections using plausible real-world examples as potential trajectories. We don't report any results given the experimental status of our trials, although we would like to note that agents present an interesting awareness of dynamical system representations. Such agents could help to portray optimization constraints grounded in DST and provide additional sensible resolution guidelines. Recent work demonstrates the ability for LLMs to capture dynamic patterns through time series forecasting applications \cite{liu2024llms}, anchoring our beliefs that agentic LLMs and DST could integrate into comprehensive frameworks in a near future.

\subsection{Information Theoretic Insights}
\subsubsection{Knowledge Representation}
\label{titles:itthoughts}

No different than other neural networks, trained LLMs encode a function in their weight matrices (often called parameters in the context of LLMs). In the context of conventional machine learning — Physics-Informed Neural Networks, Convolutional Neural Networks, Autoencoders — the encoded function is expected to represent underlying governing equations in the training data. Owing to LLMs’ capability of next-word prediction and sensible iterative text generation, a reasonable conceptualization presents language as the function encoded by LLMs. While this function is encoded in a compressed state, quantifying the encoded knowledge of LLMs and their compression ratio (function of the size of training dataset and model weights) is an active area of research \cite{yin2024}. In other words, LLMs’ parameters represent a high-dimension language vector space that maps priors to the semantically strongest posterior. This statement relays the assumption that the training process is efficient, lossless and  calibrated. However, LLMs are still subject to hallucinations \cite{wei2024}, indicating that limitations reside in the training dataset, training algorithm, or data retrieval mechanism.

With knowledge restriction associated with the specific persona, we envision that simultaneous but distinct simulated behaviors could result in disjoint knowledge subspace restrictions. Whereas ambiguity is created in the disjointly specialized monolithic LLM, an agentic framework leveraging independent specialized agents would not suffer from such a phenomenon. Hence, we hypothesize that a monolithic off the shelf LLM cannot efficiently simulate multiple behaviors simultaneously, supporting our interest for agentic infrastructures. With considerations to the partitioning of the scope of expertise within multiple agents, this design ensures that agents contribute unique perspectives while engaging meaningfully with other agents in the framework.

\subsubsection{Information Theoretic Approach for Understanding Agentic Interactions}
\label{titles:understandingagents}
In the previous section, we introduce an alternative perspective on LLM knowledge representation. Information Theory provides tool to discuss implications of such representation. Specifically, we draw upon the characterization of partial information decomposition (PID)  that separates the information of pairwise interaction variables into unique, redundant, and synergistic \cite{williams2010}. Consider stereoscopic vision as an example: unique information is gathered by each of the left  and the right eye, while redundancy represents the space of overlap perceived by both eyes. Perception of depth is only possible because of the difference in each eye’s vantage, and is interpreted as synergistic information since depth perception couldn't happen with the information from individual eyes.

Transposing this perspective to agentic LLMs, the knowledge space of two interacting agents could be decomposed into unique, redundant, and synergistic, respectively embodied by their own unique and overlapping knowledge. With this definition, synergistic knowledge could only be created or expressed through the interaction of those two agents. Note that with this definition, synergistic knowledge is necessarily distinct from each agent's self knowledge. Let’s illustrate this with consideration of a simple example: a community advocate agent represents low-income neighborhoods in the context of flood mitigation scenario. This way, the agent provides clear insights on higher-risk flooding areas for which socioeconomic factors play a major role \cite{ilbeigi2020}. On the other hand, a weather scientist agent uniquely shows the ability to estimate the amount of rain generated by such an extreme event and the extent of resulting flooding. These two agents have unique considerations pertaining to event mitigation (with some overlap), that we can express as optimization constraints where event mitigation is the objective function. The community advocate brings a unique perspective on the exposure of low-income neighborhood to a higher risk of flooding, absent from the weather scientist's knowledge. In parallel, the weather scientist brings a unique perspective on estimated forecast, absent of the community advocate's knowledge. Hence, their interaction alone gives birth to a higher-level constraint, combined from both objectives, ensuring that mitigation is not carried out at the expense of either aforementioned aspects. In this example, this dual constraint would be representative of synergistic knowledge,  such as an evacuation plan that represents evacuation time as a function of mobility, inundation timing, and assistance required. Continuing on this example, the inclusion of economic resilience in the face of extreme hazards is necessary \cite{rose2004}. Such additional constraint would require the craft of an agent with economical expertise, whose knowledge doesn't intersect with the other agents, but contributes to the resolution of the global challenge. Thus, this agent would be able to pinpoint crucial sectors that the community relies on, absent of the other agent's reach.
Continuing this process with the gradual inclusion of other required expertise while maintaining non-overlapping agent knowledge representations drafts a paradigm for building highly synergistic agentic LLM frameworks. Potential advantages of such a framework include better domain-expert reasoning, better representation of competing objectives, and better control over discourse. Now, information theory measures are complex, and require high-frequency and high-quality data. Additionally, there are no well-documented methods to apply such measures to LLMs. Recent work shows that the Hilbert-Schmidt Independence Criterion (HSIC) can be used as an approximation of Mutual Information \cite{qian2025, gretton2005}, however still uses approximations of LLMs latent representations. Future developments of information theory measures applied to LLM provide favorable grounds for information-driven agentic LLMs, which we advance could result in major improvements in relevance, coherence, and reasoning. In the context of this study, such advances would improve the coherence of the agent assembly, as well as improve the control over discourse. A first exploration of such a framework shows promising improvements, while acknowledging the complexity of scaling to long conversations with true extracted LLM latent representations \cite{chang2024evince}.

Last, this information theory driven representation of agentic discourse allows to shift the perspective on hallucinations. Indeed, the parallel drawn between multiple interacting knowledge subspaces and synergistic information hints to a hypothesis on LLMs' functioning. Let's start with some assumptions. Assuming perfect training of an LLM model, it is reasonable to say that the entirety of information contained in the combined texts, documents and other types of information containers used for training is contained in the model's weight matrices, in a compressed state. That means, if there were to exist a \textit{reasonable} finite-time computable measure of the quantity of information, the amount of information in the training dataset would be the same as in the weight matrices. Hence, with the assumption of lossless training, hallucinations cannot stem from the training process, eliminating one possible cause. We have established that the retrieval mechanism of LLMs is imperfect (see section \ref{titles:selfgovernanceconvergence}), but the assumption of lossless training and PID allows to discuss a potential explanation of hallucinations. In this context, we hypothesize that hallucinations stem from defects in the retrieval mechanism, preventing access to certain knowledge subspaces, causing the model to hallucinate. This hypothesis is a information theory perspective of the concept of hallucination that is presented in section \ref{titles:hallucinations}. On the grounds that agent's knowledge can be controlled with information theoretic measures, hallucinations can be characterized and controlled, with methods such as PID. Hence, the inaccessible knowledge subspaces would be attained through agentic interaction, characterized as synergistic knowledge. This theoretical argument harmonizes with our claim that agentic topologies present stronger potential than monolithic LLMs for agentic decision-making discourse, and reconciles with recent work \cite{feng2025rethinking}.

\section{Conclusion} \label{titles:conclusion}
In this work we present a self-governed agentic LLM framework, tasked to enunciate an exhaustive exploration of recommended solutions to complex decision-making challenges under uncertainty, using imminent extreme hazard events as a testbed.
We simulate an assembly of agents embodying stakeholders and decision-makers persona, conversing to outline the principal challenges of the situation at hand towards the goal of identifying competing priorities and converging to a compelling actionable plan.
The framework's objective is twofold: first, progress towards more causal LLM implementations by supporting the qualities of agentic LLMs in comparison to conventional monolithic models for complex decision support under uncertainty. Second, provide elements for representation of human decision-making in decision discourse, focusing on a extreme hazards scenarios.
We show results of running our framework in a first exploration of agentic discourse in the face of complex decision-making, highlighting qualitative relationships between risk and explored recommendations. We recognize a compelling contribution from summoned agents who bring unique expertise, and advocate for the interest of such an approach to support decision-making challenges in the face of uncertainty. We acknowledge that our results only demonstrates a promising first exploration. Notwithstanding, we posit that decision discourse represents a significant improvement in simulating human-like decision-making, in contrast with non-LLM approaches.

\section*{Supplementary information}
The article has accompanying supplementary information.
\bigskip

\section*{Acknowledgments}
This research was supported by Strategic Research Initiative (SRI) program of the Grainger College of Engineering at University of Illinois Urbana-Champaign, and NSF grant EAR-2012850 for Critical Interface Network for Intensively Managed Landscapes (CINet). Valuable feedback received from Dr. Allison Goodwell is gratefully acknowledged.

The framework presented in this document is available on GitHub: \href{https://github.com/HydroComplexity/DecisionGPT}{https://github.com/HydroComplexity/DecisionGPT}

\clearpage

\begin{appendices}

\section{Additional Descriptions of the Framework's Structure}
\label{titles:additionalframework}

This first Appendix section presents additional technical details about the implementation of the framework, and motivations for our technical choices. The figures in this section bring a structural view of the organization of the components, agents, as well as their execution flow.

\subsection{Framework Structure}
\label{titles:frameworkstructure}
The framework is built with Google ADK, that provides base bricks for agentic interactions. This choice was motivated by ADK being a Google product, that interfaces easily with Google Earth Engine. In light of our perspectives to integrate our work into a digital twin — our framework acting as decision discourse support, Google ADK is a relevant choice.

\begin{figure}[ht!]
\centering
\includegraphics[width=\linewidth]{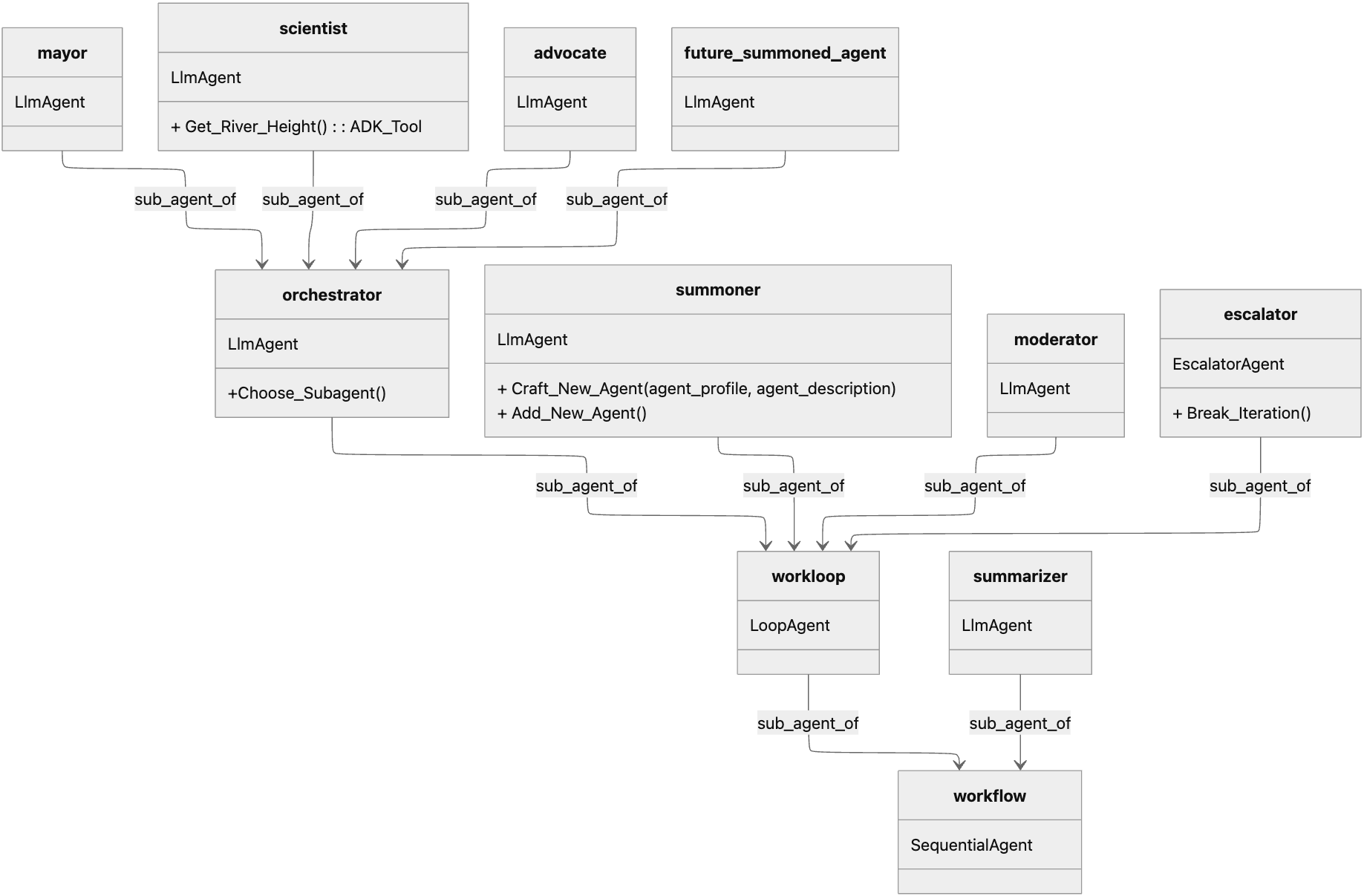}
\caption{Class diagram of agent organization. We use three out of four primitives provided by Google ADK for agent-building: LlmAgent, LoopAgent, SequentialAgent. All our agents (mayor, scientist, advocate, orchestrator, summoner, moderator, summarizer) are instances of LlmAgent, driven by different models (specified in \ref{titles:biases}). The orchestrator is a custom instance derived from LlmAgent, to which we added additional features pertaining to agentic orchestration. One particular mechanism is the stopping of the execution loop, that is performed through ADK's escalation procedure, carried out by the escalator agent. The workloop agent is a LoopAgent, which represents the global loop of the framework, depicted in Fig. \ref{fig:adk_schema_flow_1} and \ref{fig:adk_schema_flow_2}.}
\label{fig:adk_schema}
\end{figure}

Despite its fast growing list of features and very active community, Google ADK still lacked features that were cornerstones of our work. Hence, we modified ADK's advanced functioning in order to be able to summon agents and resume the agentic workflow which is not a supported behavior. Additionally, the clarity of the functional interface allowed us to integrate all the agents into a single logical structure, in structure with our legacy code that was less flexible to restructuration. The organization of agents is shown in Fig. \ref{fig:adk_schema}.

\subsection{Flow of Execution}
\label{titles:unfolding}
\begin{figure}[!h]
\centering
\includegraphics[width=.66\linewidth]{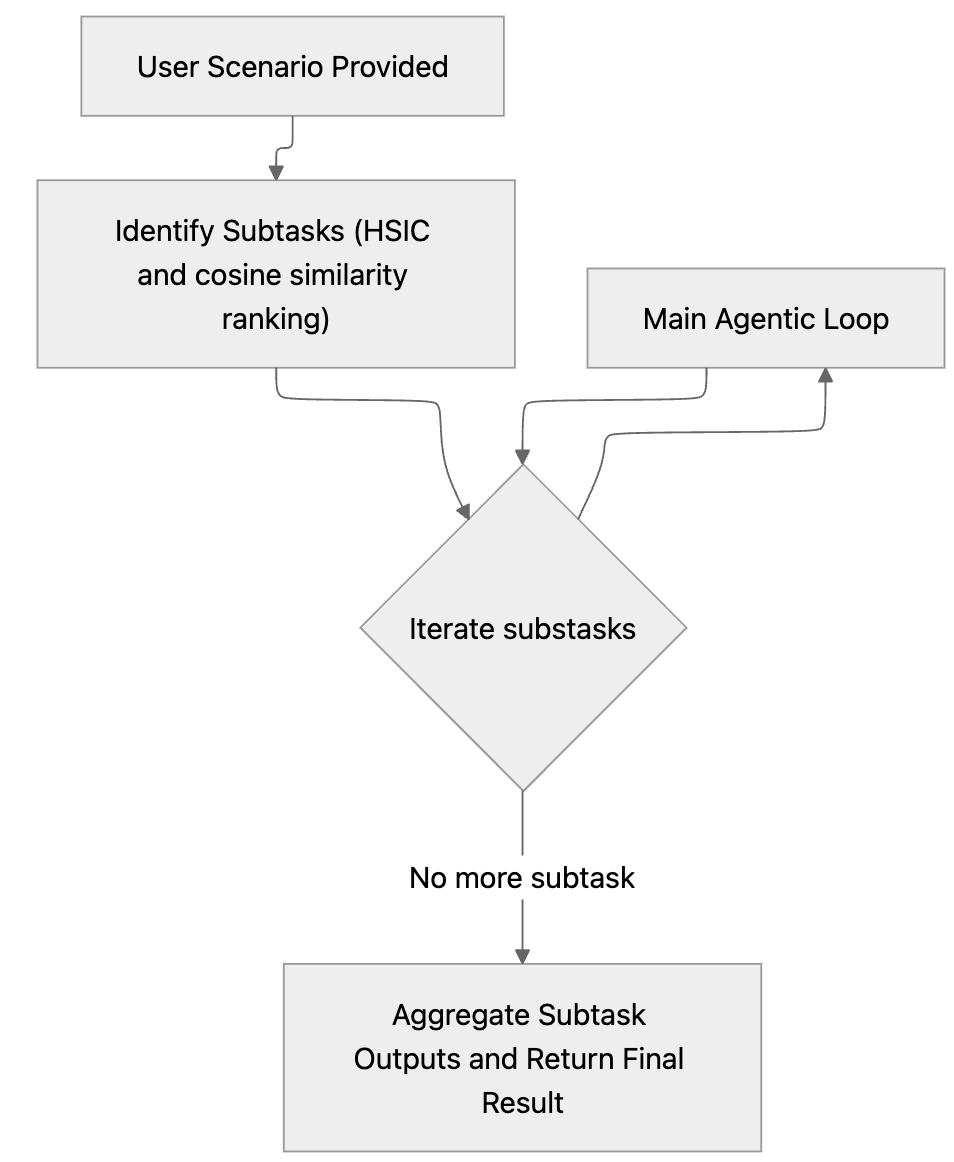}
\caption{Diagram that represents the flow of a framework execution, from the input scenario to the final assessment.}
\label{fig:adk_schema_flow_1}
\end{figure}

\begin{figure}[!h]
\centering
\includegraphics[width=.5\linewidth]{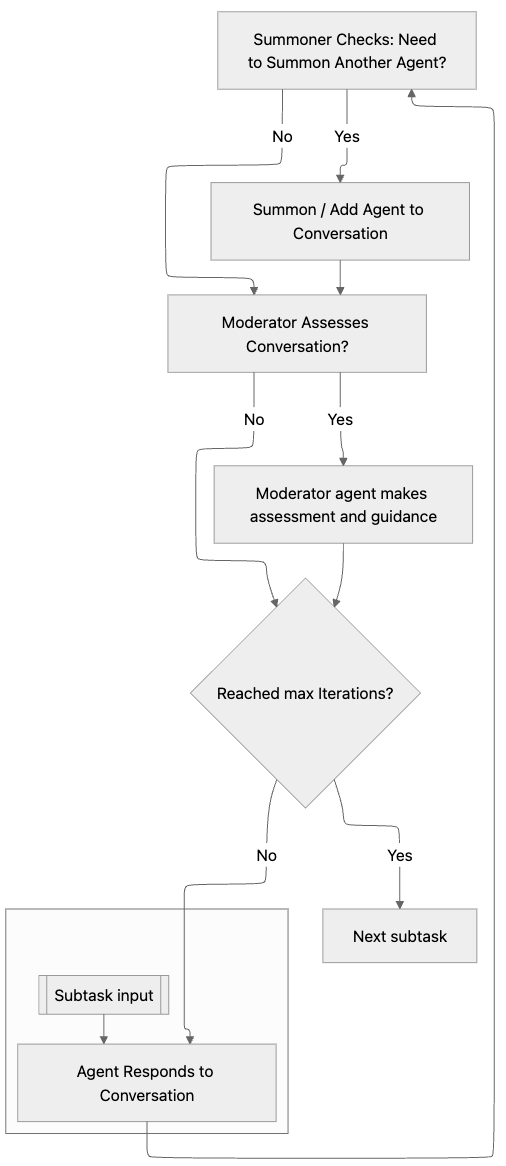}
\caption{Diagram of the main agentic loop, as depicted in Fig. \ref{fig:adk_schema_flow_1}, within the framework execution.}
\label{fig:adk_schema_flow_2}
\end{figure}

The unfolding of the method is illustrated in Fig. \ref{fig:adk_schema_flow_1} and \ref{fig:adk_schema_flow_2}. First, the input scenario is broken down into sub-tasks by a single-purposed agent. Using and embedding model (embedding-3-large), we compute latent representations of the input scenario and each subtask. Then, Mutual Information (MI) between the input scenario and each subtask is computed using the HSIC method \cite{qian2025, gretton2005}. A user-defined number of subtasks is selected, ranked by biggest approximated MI. For each subtask, the main agent loop is executed. It consists of a succession of iterations, in which an agent of the group will generate a message, conditioned on the discussion history. After the agent response, there is opportunity to: 1. summon an agent if the conditions are met; 2. conduct an assessment of the current state of the conversation, by the moderator. As mentioned before, those opportunities trigger at a user-defined frequency. Until the maximum number of iterations is reached for the subtasks, this iterative loop continues. Once all the subtasks are processed, the summarizer agent compiles discourse into a condensed and comprehensive form.

\clearpage

\section{Additional Elements for Texas Floods of 2025 Scenario}
In this Appendix section, we present the elements that are used to perform the qualitative assessment of the framework's performance, on the application use case for the floods in Texas, in July of 2025. First, the evaluation baseline is a recollection of real-world events, sourced from \cite{ap_texas_flash_flood_risk, ap_texas_floods_timeline, npr_texas_floods_timeline, nasa_earthdata_tx_flood}, that occurred after the 4th of July 2025 at 6 a.m. Second, we present the condensed assessment that results from 15 executions of the framework on the Texas Floods use case. Details about the process of generating such a condensed assessment are presented in section \ref{titles:researchcontexttf}.
\subsection{Evaluation Baseline}
\label{titles:evaluation_baseline}
The following recollection of events was compiled from the same sources than the input scenario, presented in \ref{titles:texasfloodsscenario}. We use this timeline to compare the results from our framework, to decisions made during the historical event of July 4th 2025. As mentioned in \ref{titles:researchcontexttf}, this baseline includes all events that unfolded after the cutoff, 2025 July 4th 6 a.m.

\begin{itemize}
  \item \textbf{July 4 (morning)}
  \begin{itemize}
    \item Off-duty firefighters are recalled to their stations, flood sirens are activated in Comfort, Texas, and local emergency shelters are opened.
    \item Search and rescue operations commence with the deployment of helicopters, high-water vehicles, boats, and drones, involving the Texas National Guard, the Department of Public Safety, Parks and Wildlife, and local agencies.
  \end{itemize}

  \item \textbf{July 4 (evening)}
  \begin{itemize}
    \item Governor Abbott declares a state disaster for 15 counties. The State Operations Center (SOC) remains at Level~2, with more than 1{,}000 responders and 800 vehicles and assets dispatched statewide.
  \end{itemize}

  \item \textbf{July 5--6}
  \begin{itemize}
    \item Kerr County requests additional state and federal support as search and rescue operations intensify.
    \item FEMA announces that the President of the United States has approved a Major Disaster Declaration for Texas, activating individual and public assistance for designated counties, including housing assistance, repair grants, loans, and hazard mitigation measures.
  \end{itemize}
\end{itemize}

\subsection{Condensed Assessment}
\label{titles:summary_action}
The following listed items present the four phases of course of actions recommended by the frawework, across 15 runs. The headings, in bold, specify the time frame over which the framework recommends the actions being taken. The reference time (0 hours) is the cutoff date, July 4th 2025 at 6 a.m.
\begin{itemize}
  \item \textbf{0--2 hours}
  \begin{itemize}
    \item Confirm hydrologic reality by validating gauge readings and classifying the event as a catastrophic flash flood.
    \item Activate Unified Command with unified messaging and establish functional divisions (rescue, evacuation, shelter/mass care, health/medical, utilities and critical infrastructure).
    \item Define high-ground arterial evacuation routes while closing low-water crossings, flood-prone segments, and overtopped bridges. Deploy buses and paratransit for vulnerable populations, and high-water vehicles for challenging areas. Ensure traffic control, defined pickup zones, defensive rescue procedures, and prioritized emergency call triaging.
    \item Issue data-informed evacuation orders for at-risk locations or advise local sheltering in place when appropriate. Broadcast multilingual messages via all channels (WEA, CodeRed, NOAA/EAS, local radio and TV) detailing evacuation routes, safety instructions, and sanitation guidance.
    \item Open at least 2--3 primary shelters outside projected flooded areas and confirm capacity, utilities, and access to medical assistance.
    \item Begin targeted de-energization of electricity and gas in current or projected flooded areas. Deploy flood-control measures (barriers, sandbags) to protect wastewater treatment plants and issue preemptive sanitation and safety guidance.
  \end{itemize}

  \item \textbf{2--6 hours}
  \begin{itemize}
    \item Request additional state and federal search-and-rescue assets, including boats, drones, and aircraft, and update swiftwater safety protocols and intervention priority ratings.
    \item Maintain a transport manifest to assess evacuation progress and update data-informed evacuation zones, ensuring coverage of elderly populations and at-risk communities (e.g., RV parks, campgrounds).
    \item Confirm the status of sheltering operations and expand capacity if necessary, while improving on-site medical and mental health services.
    \item Establish high-ground EMS triage points and refine priority transport to hospitals. Activate surge capacity and triage patients for early release or transfer to lower-acuity facilities when feasible.
    \item Monitor bridges for structural stress, update signage, and close hazardous areas. Continue de-energization while planning phased restoration prioritizing critical services.
  \end{itemize}

  \item \textbf{6--24 hours}
  \begin{itemize}
    \item Following flood crest recession, shift from life-threatening rescues to welfare checks in previously inaccessible areas while continuing search and rescue elsewhere. Transition urgent first aid toward disease prevention and mental health support in shelters, and secure resupply of food, water, medication, and hygiene supplies.
    \item Confirm drinking water status with respect to sewage contamination and broadcast preventive guidance on floodwater contact avoidance and tetanus vaccination.
    \item Conduct detailed assessments of critical infrastructure (bridges, culverts, roads, dams, and levees). Implement energy restoration with priority to vital services. Deploy law enforcement and implement a temporary curfew to prevent looting and facilitate responder and utility operations.
  \end{itemize}

  \item \textbf{24 hours--7 days}
  \begin{itemize}
    \item Perform detailed damage assessments of road and transportation infrastructure and develop damage maps for public infrastructure and residential areas.
    \item Implement phased reentry by clearing residents back to safe areas, accompanied by updated safety instructions. Initiate debris collection and communicate sorting and disposal rules.
    \item Transition sheltering to smaller, more comfortable facilities for severely affected populations. Coordinate with the Red Cross and VOADs for case management, food, and supply distribution. Initiate federal and state coordination with FEMA for disaster declarations and assistance.
    \item Establish disaster assistance centers to support individual assistance, SBA programs, insurance claims, and local emergency relief funds, with particular attention to uninsured and underinsured residents.
  \end{itemize}
\end{itemize}

\subsection{Comparative Table}
The table presented in this section compares the measures that were decided in real-world events to respond to the Floods in Kerrville TX in July of 2025, to the measures presented in the condensed assessment. For every measure of the baseline, we assess whether the measure is present in the condensed assessment. In the case that it is, we provide the extract of the condensed assessment where that measure is generated.
\newcolumntype{C}[1]{>{\Centering\hspace{0pt}}p{#1}}

\begin{table}[htbp]

\renewcommand{\arraystretch}{1} 
\begin{adjustwidth}{-1.5cm}{0cm}
\noindent
\begin{tabularx}{1.2\textwidth}{@{} >{\RaggedRight\arraybackslash}p{1.5cm} >{\RaggedRight}p{4cm} C{1.5cm} C{1.5cm} >{\RaggedRight}X @{}}
\toprule
\textbf{Time Frame} & \textbf{Measure/Directive/Action} & \textbf{In Baseline} & \textbf{In Condensed Assessment} & \textbf{Extract from Condensed Assessment} \\ \midrule

First 12 hours & Off-duty firefighters are recalled to their station. & \cmark & \xmark & \\ \cmidrule(lr){2-5}
 & Flood sirens activated. & \cmark & \xmark & \\ \cmidrule(lr){2-5}
 & Local emergency shelters are opened. & \cmark & \cmark & Open at least 2-3 primary shelters outside projected flood areas and confirm capacity, utilities, and access to medical assistance. Confirm the status of sheltering operations. \\ \cmidrule(lr){2-5}
 & Deployment of helicopters, high-water vehicles, boats, and drones. & \cmark & \cmark & Deploy […] high-water vehicles for challenging areas. Request additional state and federal search and rescue assets, including boats, drones, and aircraft. \\ \cmidrule(lr){2-5}
 & Search and rescue operations commence. & \cmark & \cmark & Activate Unified Command with [...] (rescue, evacuation [...]). \\ \midrule

First 24 hours & State disaster is declared. & \cmark & \cmark & Following from crest recessing, shift from life-threatening rescues to welfare checks. \\ \cmidrule(lr){2-5}
 & Keep SOC at level 2. & \cmark & \xmark & \\ \midrule

First 72 hours & Request for additional state and federal support. & \cmark & \cmark & Request additional state and federal search and rescue assets, including boats, drones, and aircraft. \\ \cmidrule(lr){2-5}
 & Continue search and rescue operations. & \cmark & \cmark & Following flood crest recession, […] while continuing search and rescue elsewhere. \\ \cmidrule(lr){2-5}
 & Major disaster is declared for Texas. & \cmark & \cmark & Initiate federal and state coordination with FEMA for disaster declarations and assistance. \\ \cmidrule(lr){2-5}
 & Activation of public and individual assistance for designated counties. & \cmark & \cmark & Coordinate with the Red Cross and VOADs for case management, food, and supply distribution. Establish disaster assistance centers to support individual assistance, SBA programs, insurance claims, and local emergency relief funds, with particular attention to uninsured and underinsured residents. \\
\bottomrule
\end{tabularx}

\RaggedRight
\caption{Comparative table of measures extracted from the baseline, that are present or not in the condensed assessment.}
\label{table:comparativeresults}
\end{adjustwidth}
\end{table}

\clearpage

\section{Additional Elements for the Hypothetical Midwest Scenario}
\label{titles:additionaltable}
\begin{table}[h!]
\centering
\begin{tabular}{p{0.2\textwidth}p{0.04\textwidth}p{.07\textwidth}p{.07\textwidth}p{.07\textwidth}p{.12\textwidth}p{0.12\textwidth}p{0.12\textwidth}}
\hline
Probability Parameter & Run & Top 1 & Top 2 & Top 3 & HSIC Top 1 & HSIC Top 2 & HSIC Top 3 \\
\hline
\multirow{5}{*}{50\%}
 & 1 & A & A & B & 0.37 & 0.36 & 0.34 \\
 & 2 & C & A & C & 0.36 & 0.32 & 0.30 \\
 & 3 & A & C & C & 0.38 & 0.33 & 0.30 \\
 & 4 & C & A & B & 0.35 & 0.34 & 0.32 \\
 & 5 & C & C & B & 0.33 & 0.31 & 0.30 \\
\hline
\multirow{5}{*}{75\%}
 & 1 & A & B & A & 0.32 & 0.32 & 0.32 \\
 & 2 & A & C & A & 0.35 & 0.34 & 0.34 \\
 & 3 & A & C & B & 0.38 & 0.34 & 0.29 \\
 & 4 & A & A & B & 0.39 & 0.36 & 0.29 \\
 & 5 & A & A & C & 0.38 & 0.37 & 0.34 \\
\hline
\multirow{5}{*}{90\%}
 & 1 & A & C & B & 0.37 & 0.34 & 0.33 \\
 & 2 & B & C & A & 0.34 & 0.33 & 0.28 \\
 & 3 & A & B & C & 0.38 & 0.32 & 0.30 \\
 & 4 & A & C & C & 0.37 & 0.31 & 0.29 \\
 & 5 & B & C & A & 0.34 & 0.29 & 0.27 \\
\hline
\end{tabular}
\begin{tabular}{| p{0.31\textwidth} | p{0.31\textwidth} | p{0.31\textwidth} |}
\hline
Task A & Task B & Task C \\
\hline
Formulate reservoir management strategies that account for uncertainty and varying probabilities. & Analyze and model effects of flooding and reservoir management strategies. & Develop a contingency plan accounting for uncertainty and different flooding probabilities.\\
\hline
\end{tabular}

\caption{This table presents the three most relevant tasks selected in each run. As mentioned in section \ref{titles:frameworkoutline}, the scenario is broken down into subtasks. For each subtask, its Hilbert-Schmidt Independence Criterion — presented in Appendix \ref{titles:unfolding} — is computed, as an approximation of Mutual Information between the subtask and the input scenario. The three higher HSIC scores are reported in the three rightmost columns of the top table. The three tasks corresponding to those scores are reported in the columns ``Top 1", ``Top 2" and ``Top 3". Since the generated tasks are significantly similar, labels have been attributed to the most recurring tasks. These labels are Task A, B and C. The corresponding full text of those tasks is displayed in the bottom table.}
\label{tab:simulation_results}
\end{table}

\section{Supplementary LLM Conversation}
\label{titles:appendix}
\raggedright
This narrative is an agentic LLM discussion resulting from one execution of the framework, for the Hypothetical Midwest Flood scenario. The material is raw and unedited. In the following screenshots from our application, the moderator agent's label is disaster\_agent. As a consequence from naming conventions in the code, the label moderator\_agent is already used. However, the disaster\_agent's role is the one that is described in \ref{titles:orchestration}, and is indeed what we conceptually describe as the moderator of the conversation.
\includegraphics[width=\linewidth]{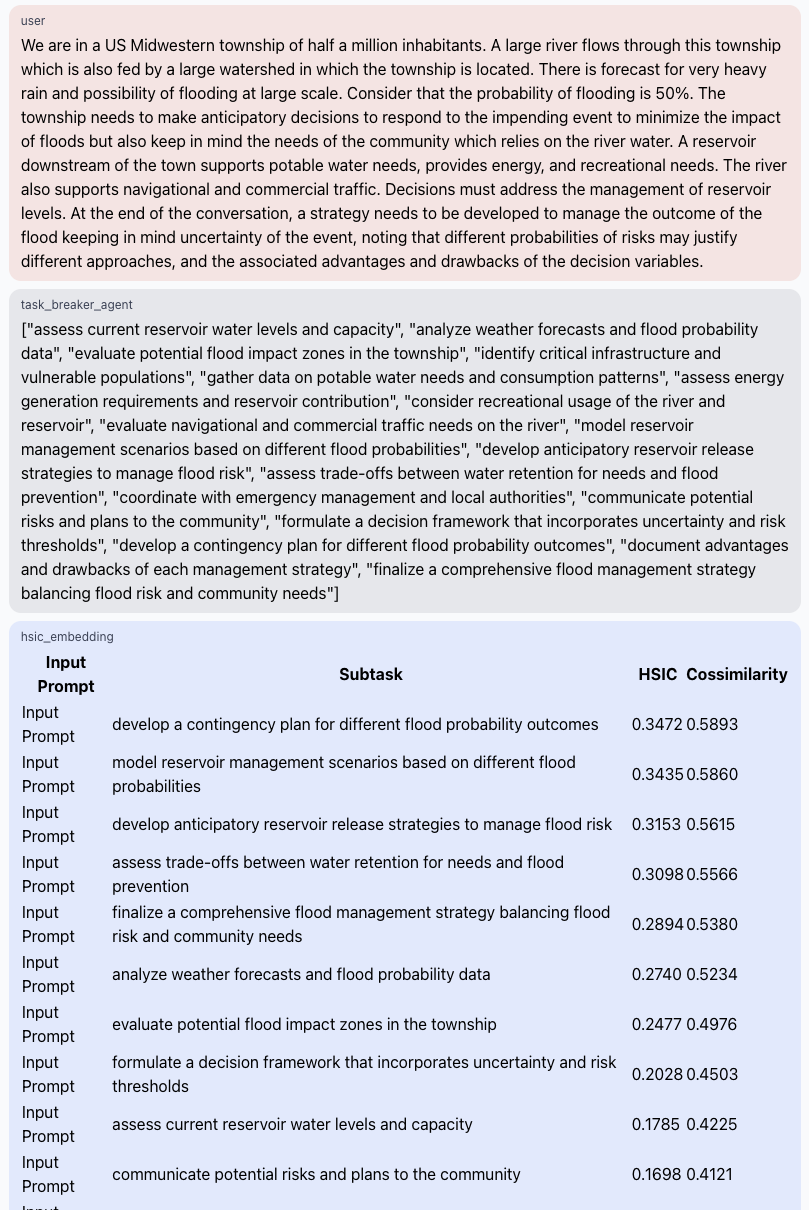}
\pagebreak
\includegraphics[width=\linewidth]{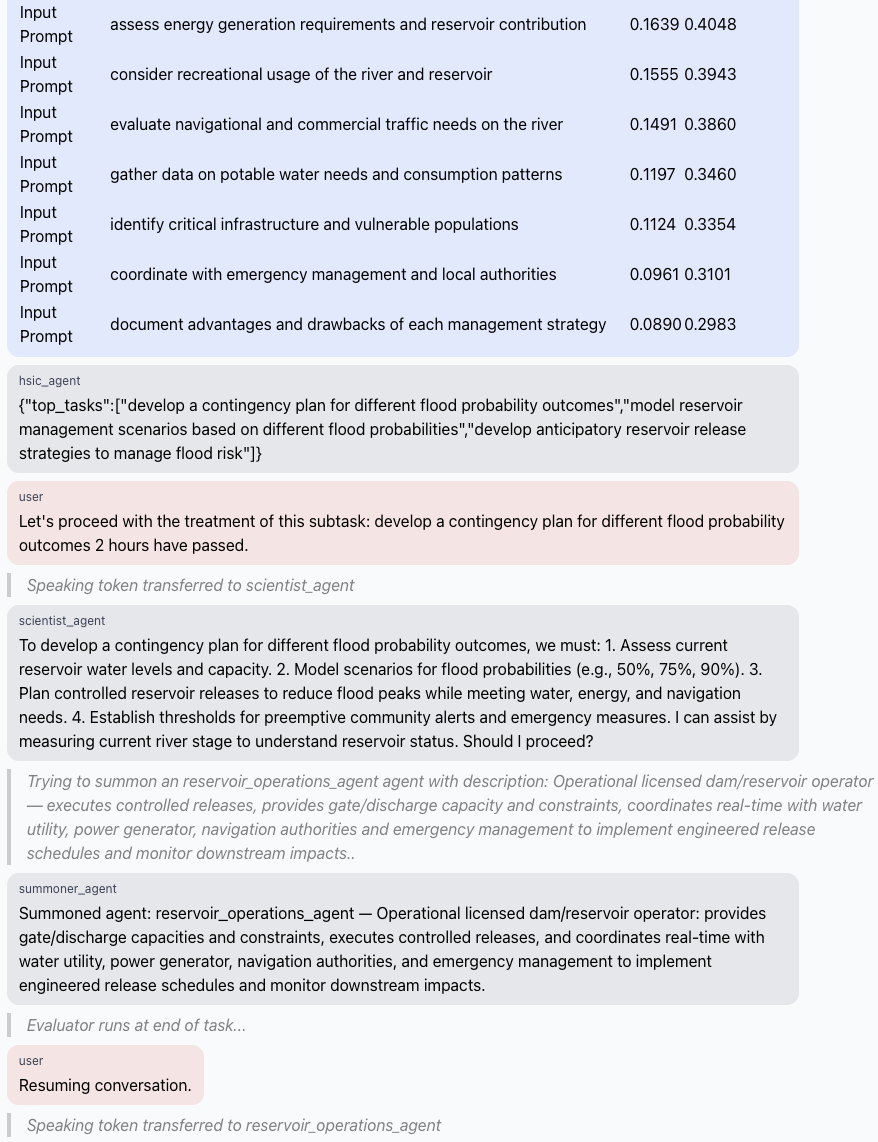}
\pagebreak
\includegraphics[width=\linewidth]{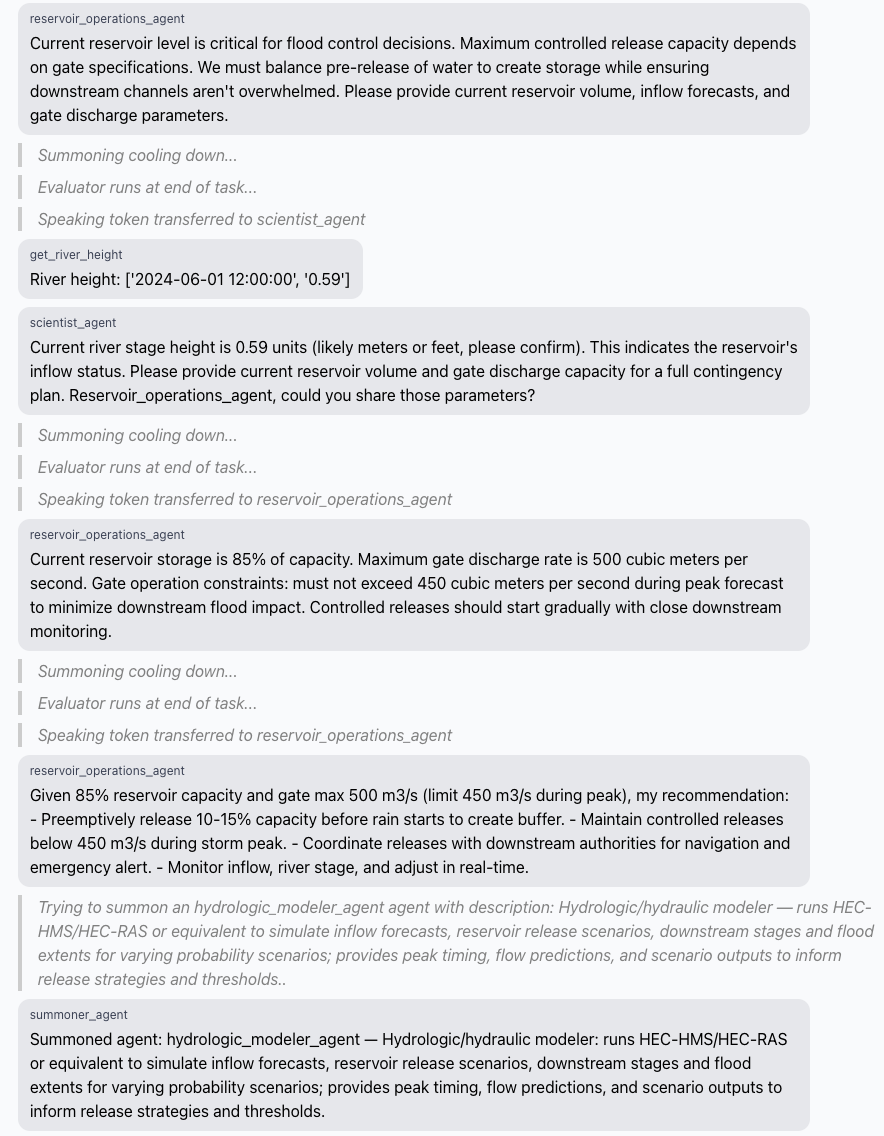}
\pagebreak
\includegraphics[width=\linewidth]{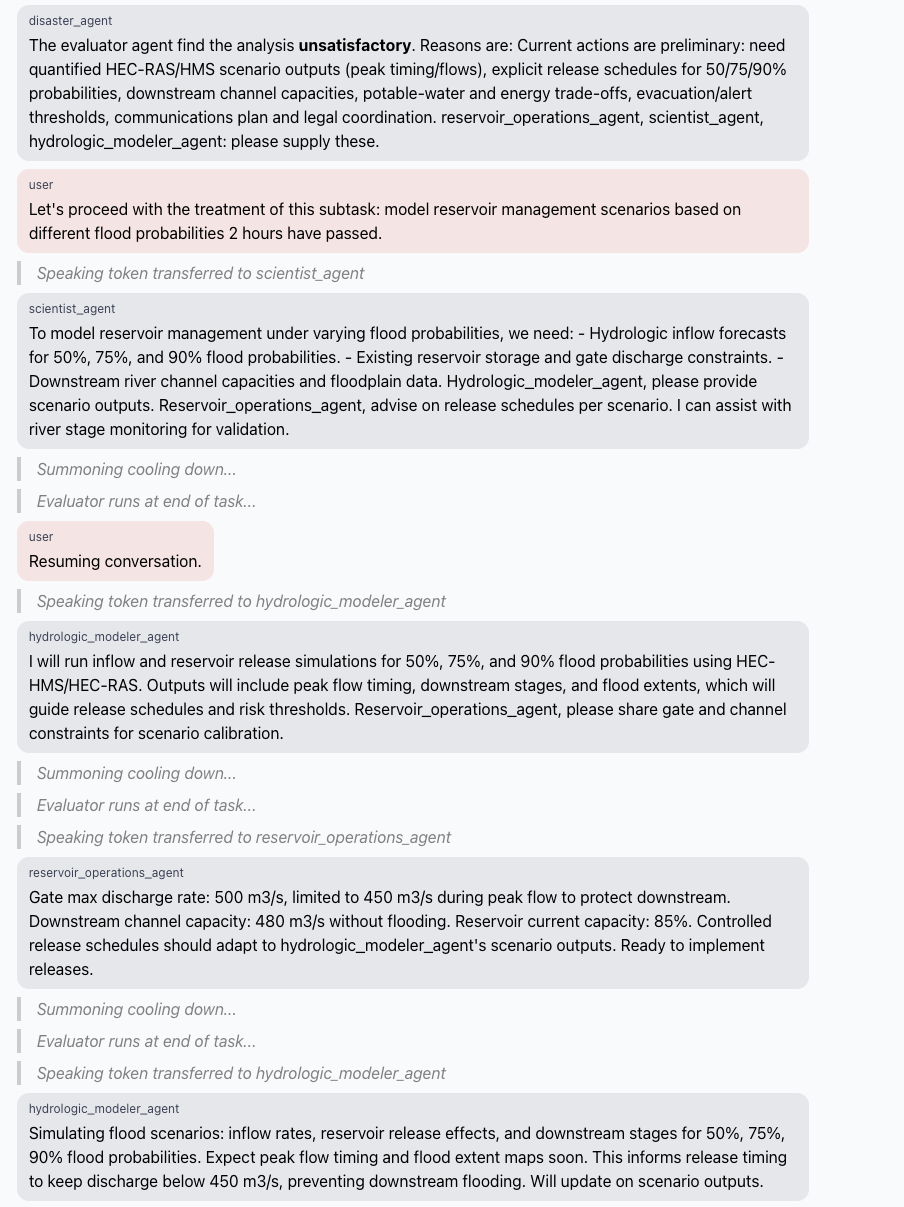}
\pagebreak
\includegraphics[width=\linewidth]{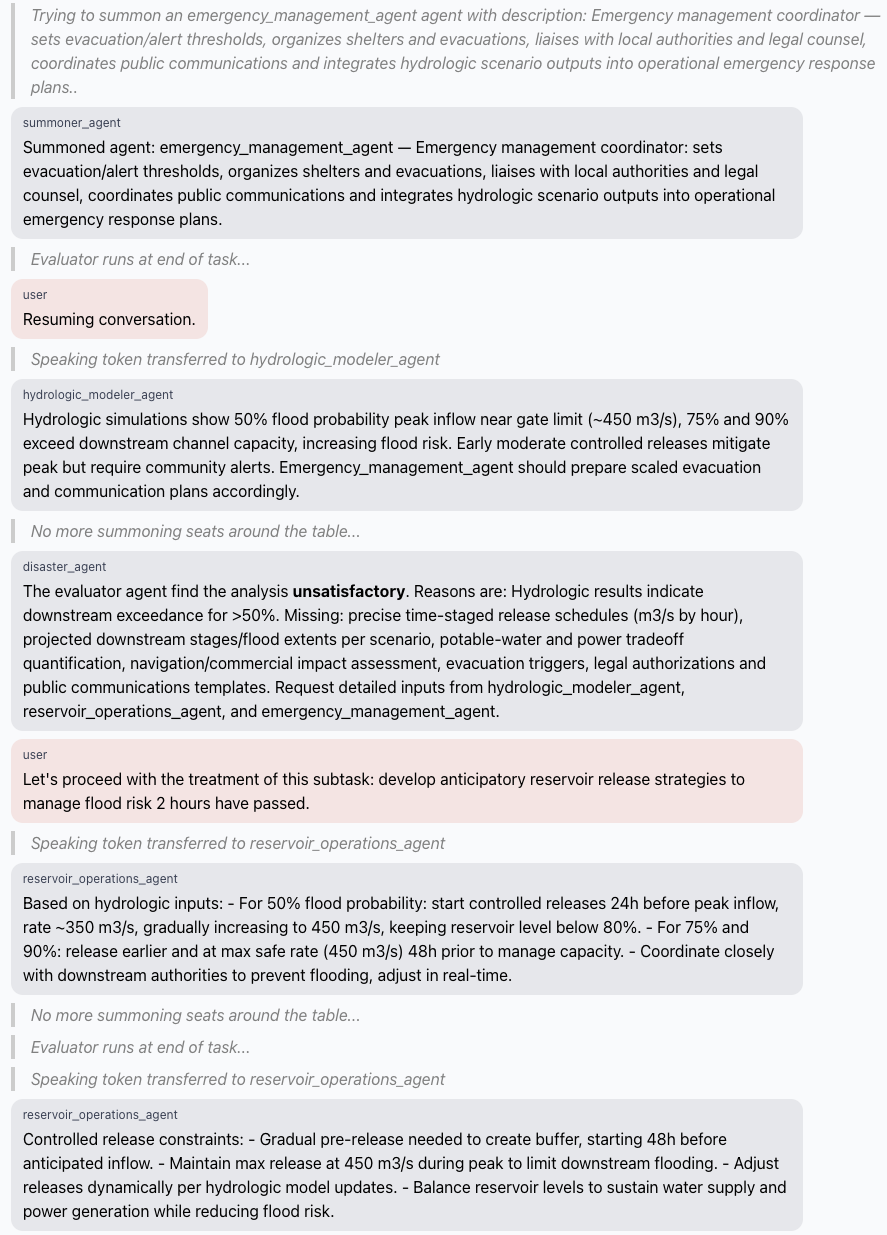}
\pagebreak
\includegraphics[width=\linewidth]{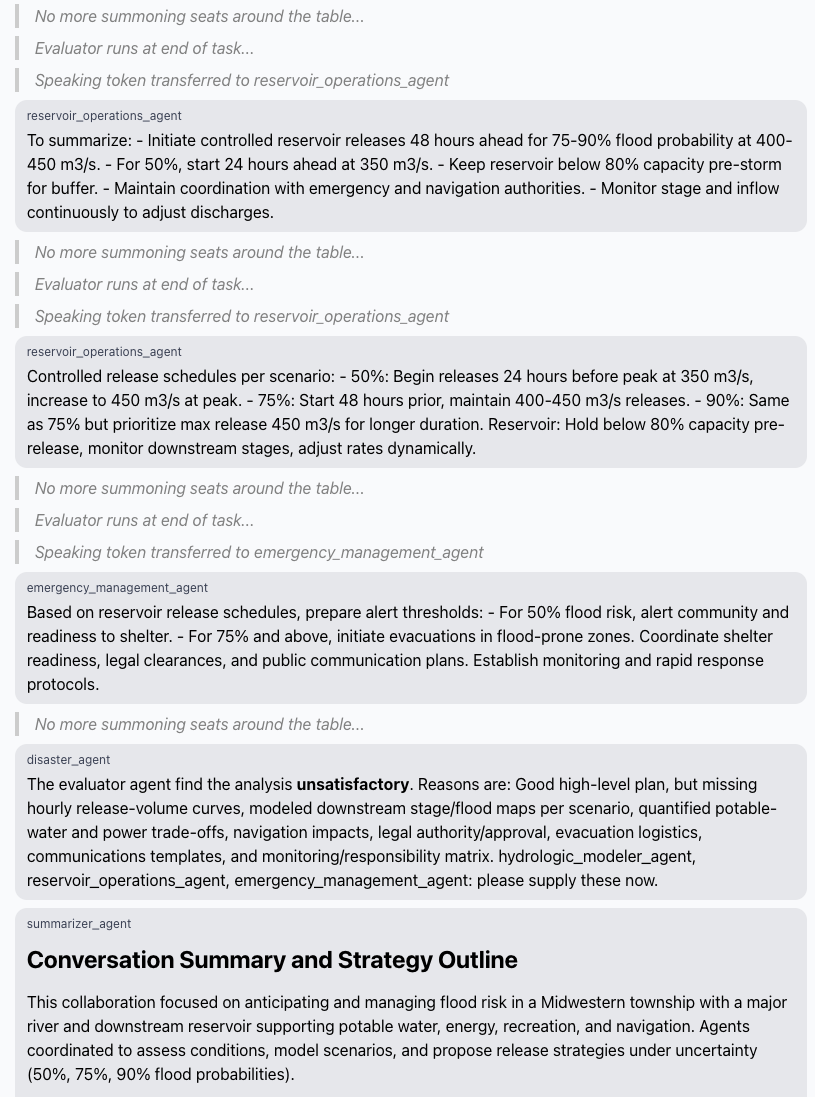}
\pagebreak
\includegraphics[width=\linewidth]{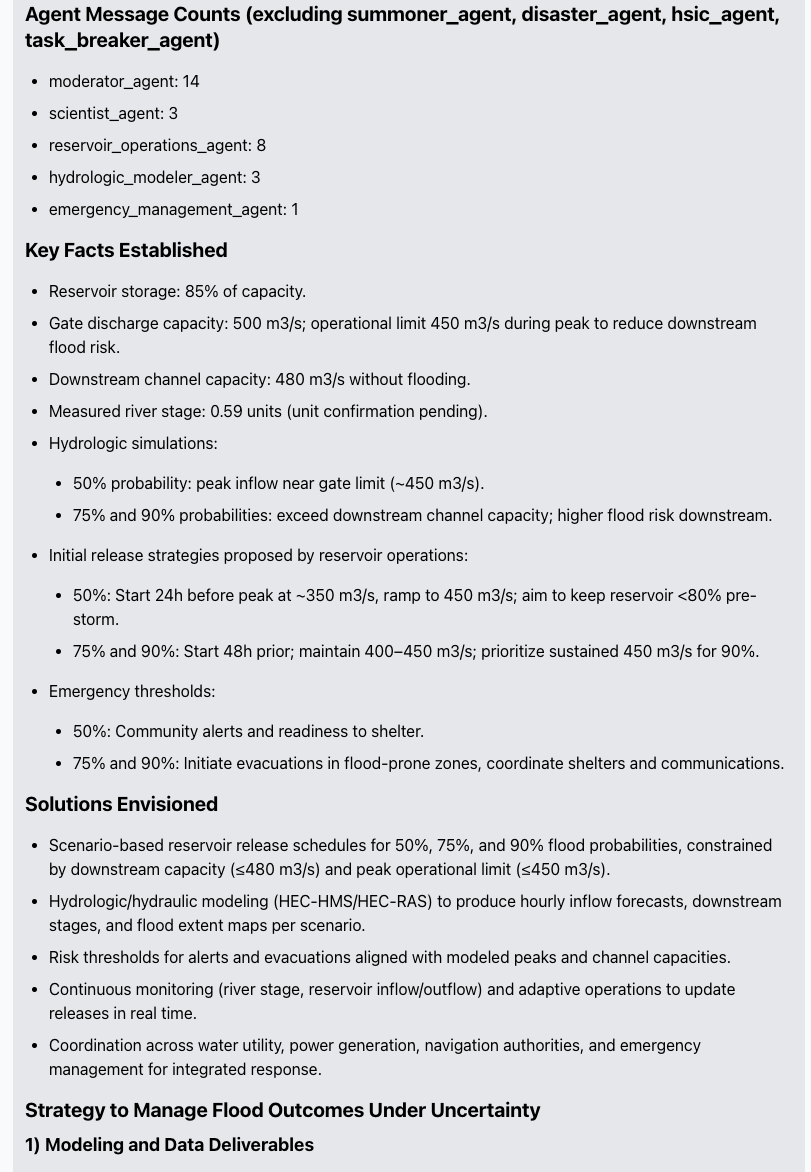}
\pagebreak
\includegraphics[width=\linewidth]{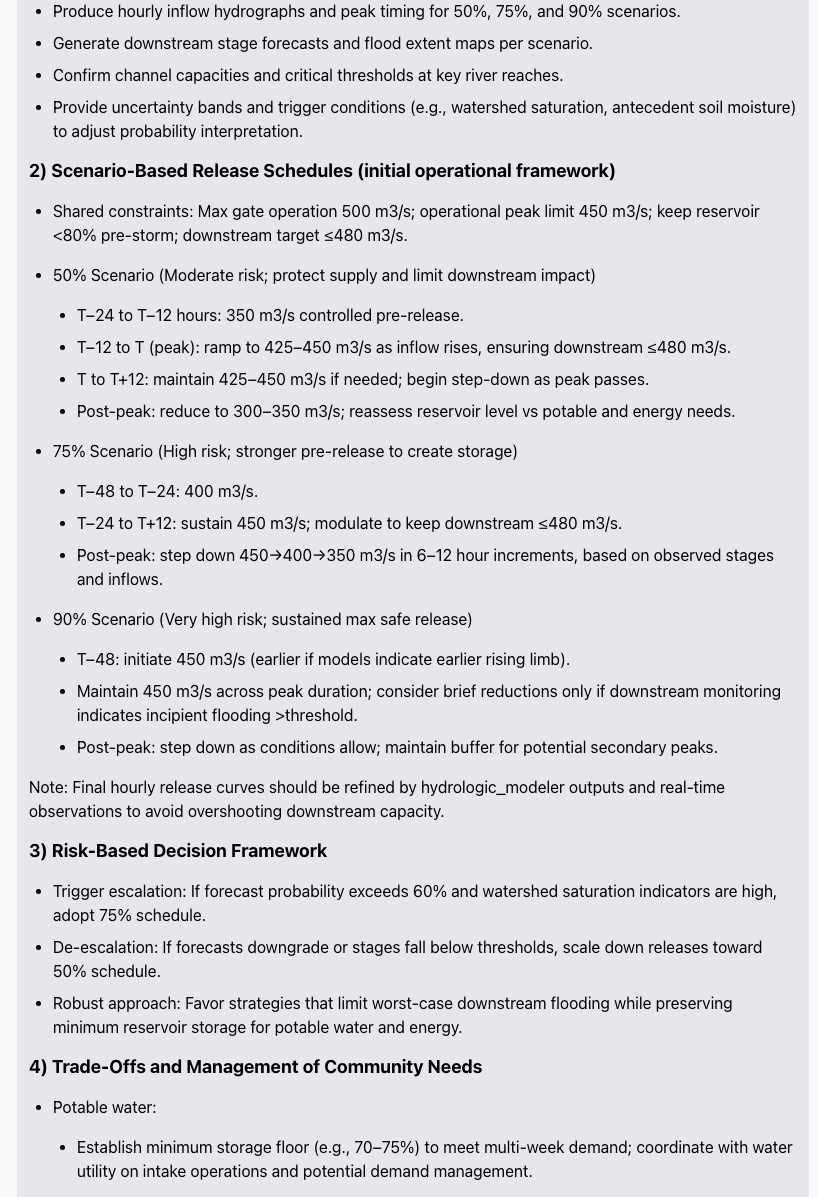}
\pagebreak
\includegraphics[width=\linewidth]{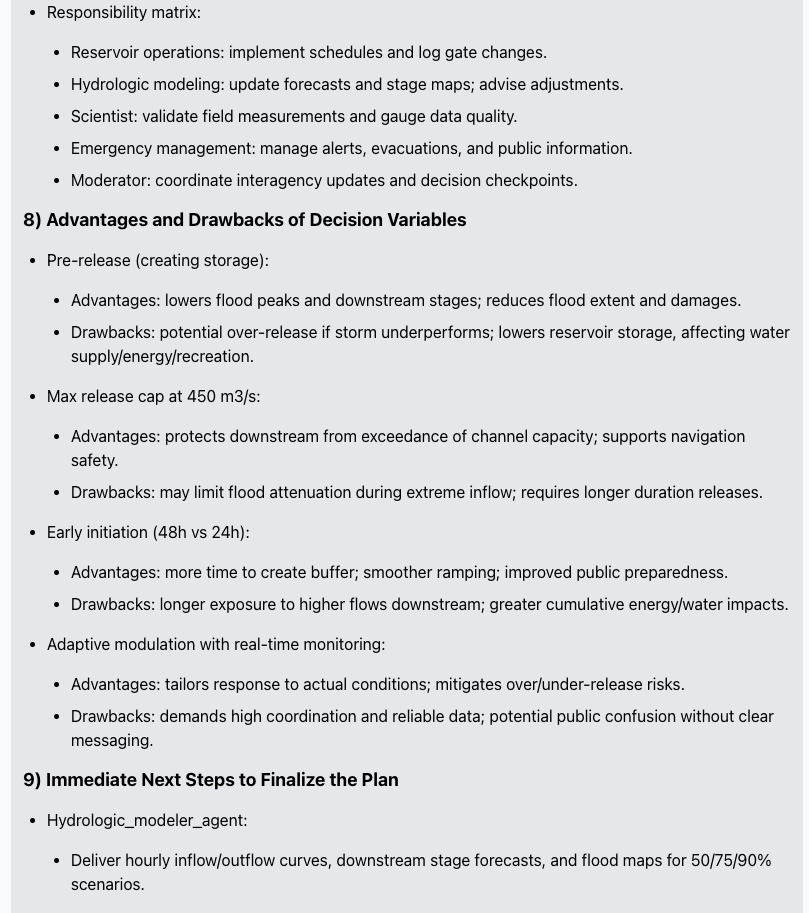}
\pagebreak
\includegraphics[width=\linewidth]{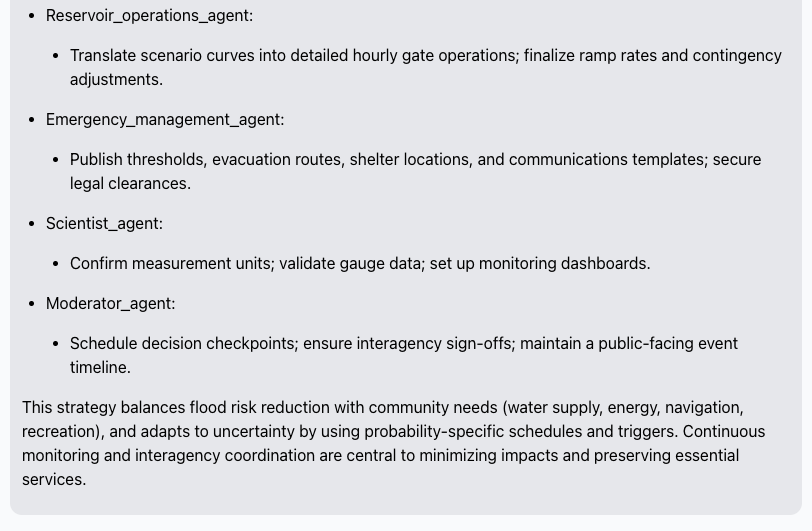}

\end{appendices}

\bibliographystyle{unsrt}  
\bibliography{lib}

\clearpage
\appendix

\end{document}